\documentclass[conference]{IEEEtran}
\IEEEoverridecommandlockouts
\usepackage{cite}
\usepackage{amsmath,amssymb,amsfonts}
\usepackage{algorithmic}
\usepackage{graphicx}
\usepackage{textcomp}
\usepackage{xcolor}
\def\BibTeX{{\rm B\kern-.05em{\sc i\kern-.025em b}\kern-.08em
    T\kern-.1667em\lower.7ex\hbox{E}\kern-.125emX}}
    
\usepackage{amsmath,amsfonts}

\usepackage{times}
\usepackage{url}
\usepackage{hyperref}
\usepackage[utf8]{inputenc}
\usepackage[small]{caption}
\usepackage{graphicx}
\usepackage{amsmath}
\usepackage{booktabs}
\usepackage{enumitem}

\usepackage{algorithm}
\usepackage{algorithmic}
\usepackage{mathrsfs} 

\usepackage{nicefrac}       % compact symbols for 1/2, etc.
\usepackage{microtype}      % microtypography
\usepackage{graphicx}
\usepackage{graphics}
\usepackage{rotating}
\usepackage{mathrsfs} 
\usepackage{array,multirow,multicol}
\usepackage{lscape} 
\usepackage{array}
\usepackage{multicol}
\usepackage{xcolor}
\usepackage{colortbl}
\usepackage{todonotes}
\graphicspath{{figs/}}
\usepackage[super]{nth}

\usepackage{algorithmic}
\usepackage{array}
\usepackage[caption=false,font=normalsize,labelfont=sf,textfont=sf]{subfig}
\usepackage{textcomp}
\usepackage{stfloats}
\usepackage{url}
\usepackage{verbatim}
\usepackage{graphicx}

\usepackage{authblk}
\usepackage{soul}
\soulregister\cite7
\soulregister\ref7
\soulregister\pageref7

\begin{document}

\title{Uncertainty Quantification for Traffic Forecasting: A Unified Approach}

%\author{Weizhu~Qian, Dalin~Zhang, Yan~Zhao, and
%        James~J.Q.~Yu,~\IEEEmembership{Senior~Member,~IEEE},
%}

%\author[1]{Weizhu~Qian}
%\author[1]{Dalin~Zhang}
%\author[1]{Yan~Zhao}
%\author[2]{Kai~Zheng}
%\author[3]{James~J.Q.~Yu
%%,~\IEEEmembership{Senior~Member,~IEEE}
%}
%\affil[1]{Department of Computer Science, Aalborg University}
%\affil[2]{University of Electronic Science and Technology of China, China}
%\affil[3]{Southern University of Science and Technology}

\author{Weizhu~Qian\textsuperscript{1},
Dalin~Zhang\textsuperscript{1},
Yan~Zhao\textsuperscript{1}, 
Kai~Zheng\textsuperscript{2},  
James~J.Q.~Yu\textsuperscript{3}
 \\
 \textsuperscript{1}Aalborg University, Denmark\\
 \textsuperscript{2}University of Electronic Science and Technology of China, China\\
 \textsuperscript{3}Southern University of Science and Technology, China\\
 \{wqian, dalinz, yanz\}@cs.aau.dk, zhengkai@uestc.edu.cn, yujq3@sustech.edu.cn\\
}

\maketitle

\begin{abstract}
Uncertainty is an essential consideration for time series forecasting tasks.   
In this work, we specifically focus on quantifying the uncertainty of traffic forecasting. %the prediction 
%\task[(06-08) Readers may not know why a model can make uncertainty. Change?]{made by} the spatio-temporal model for traffic forecasting. 
To achieve this, we develop Deep Spatio-Temporal Uncertainty Quantification (DeepSTUQ), which can estimate both aleatoric and epistemic uncertainty. 
We first leverage a spatio-temporal model to model the complex spatio-temporal correlations of traffic data.  
Subsequently, two independent sub-neural networks maximizing the heterogeneous log-likelihood are developed to estimate aleatoric uncertainty.
For estimating epistemic uncertainty, we combine the merits of variational inference and deep ensembling by integrating the Monte Carlo dropout and the Adaptive Weight Averaging re-training methods, respectively. 
Finally, we propose a post-processing calibration approach based on Temperature Scaling, which improves the model's generalization ability to estimate uncertainty. 
Extensive experiments are conducted on four public datasets, and the empirical results suggest that the proposed method outperforms state-of-the-art methods in terms of both point prediction and uncertainty quantification.  
\end{abstract}


%%==================
\section{Introduction}
\label{sec:introduction}
%\todo[inline]{Background}
Traffic forecasting is one of the essential elements in modern Intelligent Transportation Systems (ITS). 
The predicted data, including but not limited to traffic flow, speed, and volume, can help municipalities manage urban transportation more efficiently.
%In recent years, deep learning techniques have greatly influenced the field of ITS~\cite{veres2019deep}. 
%For instance, Deep Reinforcement Learning can be applied to traffic control~\cite{haydari2020deep}. 
In terms of traffic forecasting, the road segments in a road network interact with each other spatially and the current state of a road segment depends on previous states, which results in complicated spatio-temporal correlations.        
Modelling the spatial-temporal correlations of traffic data is non-trivial \cite{li2018diffusion,yu2018spatio,wu2019graph,song2020spatial,bai2020adaptive,guo2019attention,zheng2020gman}.
%\task[(06-08) Better provide a bunch of references here.].  

% Spatial-temporal model
Thanks to the recent advances of deep learning techniques, a number of deep learning-based spatio-temporal models have been proposed in the field of traffic forecasting~\cite{veres2019deep},~\cite{wang2020deep}. 
Since the topology of a typical road network can be described by a graph in which each node represents a sensor and each edge represents a road segment, the spatial dependency of traffic data can be naturally extracted by Graph Neural Networks (GNNs)~\cite{jiang2021graph}.  
Correspondingly, the temporal dependency of traffic data can be modelled by Convolutional Neural Networks (CNNs), Recurrent Neural Networks (RNNs), or their numerous variants \cite{li2018diffusion,wu2019graph,song2020spatial,bai2020adaptive}.
%\task[(06-08) References here.]{}.   

%\todo[inline]{Motivation}
Despite the fact that existing methods regarding traffic forecasting have been shown successful~\cite{wang2020deep}, most of them only provide deterministic traffic prediction without quantifying uncertainty\,---\,a critical component in traffic data.
%\task[(06-08) A bit strange. Point prediction can also consider uncertainty. This sentence looks like not.]{point traffic prediction without considering uncertainty}. 
We argue that only point prediction for traffic forecasting is not always sufficient, i.e., the knowledge of the forecasting uncertainty can also be imperative to the management of transportation systems, especially in emergency situations, e.g., route planning for rescuing vehicles and ambulances~\cite{wang2021forecasting}.
%\task[(06-08) Reference]{}. 
For this reason, we aim to attain both future traffic forecasting and its corresponding uncertainty in this paper.
More specifically, the research goal includes the estimation of both epistemic and aleatoric uncertainties, which refer to model uncertainty and data uncertainty, respectively.
Aleatoric uncertainty can be obtained by two independent neural networks by estimating means and variances, respectively \cite{nix1994estimating}. 
As for epistemic uncertainty, both variational inference and ensembling are possible solutions. 
However, these two types of approaches both have their own limitations.
Variational approaches, e.g., Bayesian Neural Networks (BNNs) \cite{jospin2020hands}, are prone to modal collapse \cite{gawlikowski2021survey}.
Deep ensembling is capable of finding multiple local minimums by training a set of deterministic models, but the prediction of each trained deterministic model lacks diversity \cite{gawlikowski2021survey}.
To circumvent this problem, it needs to find a set of local minimums/solutions with certain amount of diversities.

%\task[(06-08) Reference]{}.
% \fix[]{}{One sentence to summarize the current gap. Or, ``challenge'' this paper aims to solve.} 

To this end, we develop an epistemic uncertainty quantification method %\task[(07-05) Sounds like combining the two existing approaches. Though in fact it is, need to make it more novel. This is the first sentence of a paragraph]{combing the advantages of variational inference and deep ensembling}.
having the merits of both variational inference and deep ensembling.
Specifically, to avoid training and storing multiple models for ensembling, we devise a new re-training method based on Stochastic Weight Averaging~\cite{izmailov2018averaging}.
In addition, a post-processing calibration method based on Temperature Scaling \cite{guo2017calibration} is proposed to improve the uncertainty quantification performance.  
Finally, a unified uncertainty quantification approach called Deep Spatio-Temporal Uncertainty Quantification
(DeepSTUQ) is formulated for both epistemic and aleatoric uncertainty estimation. 
Compared to existing approaches, DeepSTUQ has the following advantages: 1) DeepSTUQ can predict future traffic while provide both epistemic and aleatoric prediction uncertainty, and 2) DeepSTUQ requires training only one single model, which as a result, is fast-training, low-memory-footprint, and fast-inferring.
%\task[(06-08) Explain why require one single model is good.]{}.

%Compared to existing approaches, DeepSTUQ has the following advantages: 1) compare to existing probabilistic approaches, DeepSTUQ can predict future traffic while provide both epistemic and aleatoric prediction uncertainty, and 2) compare to existing uncertainty quantification approaches, DeepSTUQ \weizhu{requires training only one single model, which as a result, is fast-training, low-memory cost, and fast-inferring.}
%\task[(06-08) Explain why require one single model is good.]{}.
%

The contributions of this paper can be summarized as follows:
\begin{itemize}
    \item We propose a unified method that can estimate both the epistemic and aleatoric uncertainties in traffic forecasting.
    %\task[(06-08) Only flow?]{flow forecasting};
    \item We propose a novel training approach combining the advantages of variational inference and deep ensembles for epistemic uncertainty quantification.
    \item We propose a post-processing model calibration method that further improves the performance of uncertainty quantification.
    \item Extensive experiments are conducted on four public datasets, and the obtained results show that DeepSTUQ surpasses state-of-the-art methods in terms of both point prediction and uncertainty quantification.
\end{itemize}

%The remainder of the paper is organized as follows. 
%Section~\ref{sec:problem_statement} describes the problem we want to solve in this work. 
%Section~\ref{sec:method} introduces the proposed method in detail.
%Section~\ref{sec:experiment} demonstrates the experiments and results.
%The conclusions and perspectives are given in Section~\ref{sec:conclusion}. 

%%==================
\section{Related Work}
%In this section, we review the state-of-the-art methods regarding spatio-temporal based traffic prediction and uncertainty quantification.  
Our work relates to traffic forecasting regarding its application and uncertainty quantification regarding the methodology. 
Thus, we review the state-of-the-art methods from these two aspects in this section. 

%\todo[inline]{A small paragraph briefly introduce the logic flow of this section.}

\subsection{Spatio-Temporal Traffic Forecasting}
Traffic data can be regarded as multivariate time series. Hence, for traffic forecasting tasks, both spatial and temporal correlation are critical data features to learn from. 
In terms of spatial correlations, Graph Neural Network, such as Graph Convolutional Networks (GCNs)~\cite{kipf2016semi}, ChebNet~\cite{defferrard2016convolutional}, and Graph Attention Networks (GATs)~\cite{velivckovic2018graph} have become the \textit{de facto} deep learning techniques. 
As for temporal dependency, deep architectures like Gated Recurrent Networks (GRUs) \cite{cho2014learning}, Gated Convolutional Neural Networks (GCNNs) \cite{dauphin2017language}, and WaveNet \cite{van2016wavenet} have been widely applied to traffic prediction.
Base on these two types of methods, a number of deep spatio-temporal models have been proposed in the context, such as Diffusion Convolutional Recurrent Neural Network (DCRNN)~\cite{li2018diffusion}, Temporal Graph Convolutional Network (T-GCN)~\cite{zhao2019t}, Spatio-Temporal Graph Convolutional Networks (ST-GCN)~\cite{yu2018spatio}, and GraphWaveNet~\cite{wu2019graph}.
%\task[(06-08) One sentence on their merits and bads.]{}
Those methods are capable of learning spatio-temporal correlations but fail to capture multi-scale or hierarchical dependency.

More recently, Li et al.~\cite{li2021spatial} proposed Spatial-Temporal Fusion Graph Neural Network (STFGNN), in which a spatial-temporal fusion graph module and a gated dilated CNN module were used to capture local and global correlations simultaneously.
Zheng et al.~\cite{zhang2021traffic} proposed Spatial-Temporal Graph Diffusion Network (ST-GDN) that adopted a hierarchical graph neural network architecture and a multi-scale attention network to learn spatial dependency from local-global perspectives and multi-level temporal dynamics, respectively.
Additionally, Attention Mechanism has been applied to address this issue as well. 
For instance, Attention-based Spatial-Temporal Graph Convolutional Network (ASTGCN)~\cite{guo2019attention} uses spatial attention and temporal attention to model the spatial patterns and dynamic temporal correlations, respectively.
%\task[(06-08) The reason of introducing these two methods with their implementations is unclear. Please summarize.]{}

%Attention
%\task[(06-08) Very loosely correlated to the two paragraphs above. Maybe need a stronger structure in this subsection.]{Attention Mechanism} has been applied to 
%Additionally,
%Attention Mechanism has been applied to spatio-temporal traffic forecasting as well. 
%For instance, Attention-based Spatial-Temporal Graph Convolutional Network (ASTGCN)~\cite{guo2019attention} and Graph Multi-attention Network (GMAN)~\cite{zheng2020gman} apply spatial attention and temporal attention to model the spatial and dynamic temporal correlations, respectively. 
% graph structure learning

%\task[(07-05) Break into smaller ones. Grammatical error.]{
Nevertheless, in a practical real-world case where the knowledge of a graph is missing,   
the physical road connectivity may not necessarily represent the real data correlation in a graph.
It is therefore beneficial to learn the graph structure from the data.
%}
%\task[(06-08) This statement is weak. Rephrase to the following idea: physical road connectivity may not represent data correlation in a graph. So we need to learn.]{we have to learn the graph structure from data}. 
To this end, methods such as Multivariate Time Series Forecasting with Graph Neural Network (MTGNN)~\cite{wu2020connecting} and Adaptive Graph Convolutional Recurrent Network (AGCRN)~\cite{bai2020adaptive} can learn the unknown adjacency matrix in a data-driven manner and consequently improve the prediction performance.
However, all those aforementioned methods only focus on providing point estimation without computing prediction intervals.
%\task[(07-05) Sounds not-so-professional. ``Confidence''? Weizhu: prediction interval is the right term here, it is more general than confidence.]{prediction intervals}.

%\task[(06-08) Again, not quite correlated to what you want to express in the literature review.]{}

%\todo[inline]{The above ST approaches are a bit verbose. Try to shorten by 1/3 column.}

%%==================
\subsection{Uncertainty Quantification}
%Uncertainty quantification has recently been a quite popular research ~\cite{abdar2021review},~\cite{gawlikowski2021survey}. 
%\task[(05-09) This looks harsh to me. The paper aims at T-ITS, and it's possible that non-deep learning reviewers review this paper. Possible to change a word?]{in deep learning}
Uncertainty quantification has recently been an actively researched area and widely applied to solve various real-world problems~\cite{abdar2021review},~\cite{gawlikowski2021survey}. 
In general, uncertainty can be classified into two categories, namely, aleatoric and epistemic. 

Aleatoric uncertainty refers to the data uncertainty caused by noise or intrinsic randomness of processes, which is irreducible but can be computed via predictive means and variances~\cite{kendall2017uncertainties} using negative log-Gaussian likelihood as the loss function.
%Epistemic uncertainty refers to model uncertainty, which is caused by data sparsity or lack of \task[(06-08) What is ``model knowledge''?]{model knowledge}. 
%Aleatoric uncertainty refers to data uncertainty, which is caused by noise or intrinsic randomness of processes. 
%The latter is \task[(06-08) What about the former?]{irreducible but can be computed} via predictive means and variances~\cite{kendall2017uncertainties} using negative log-Gaussian likelihood as the loss function.  
% Variational inference
%\task[(06-08) 1. Don't split into two paragraphs. 2. The order of A and E should be consistent with the description above.]{Epistemic uncertainty is reducible} (e.g., when training data increases) and can be modelled using Bayesian or ensembling approaches. 
Epistemic uncertainty refers to model uncertainty caused by data sparsity or lack of knowledge, which is learnable and reducible.
A widely-used method for estimating epistemic uncertainty is Bayesian Neural Networks (BNNs)~\cite{jospin2020hands}, in which 
%\task[(06-08) Grammatical error.]
a Gaussian distribution is imposed on each weight to generate model uncertainty. 
However, a typical BNN 
%\task[(06-08) Please explain so what.]
doubles the number of model parameters and requires to compute the Kullback-Leibler (KL) divergence explicitly~\cite{blundell2015weight}, which raises the model complexity and slows down the training process.
Alternatively, a simple approach called Monte Carlo (MC) dropout~\cite{gal2016dropout} performs Bayesian approximation by turning on dropout at both training and test time as opposed to standard dropout~\cite{srivastava2014dropout}.
%Alternatively, a simple approach to perform Bayesian approximation is to turn on Dropout~\cite{srivastava2014dropout} \weizhu{at both training and test time}, which is \task[(06-08) Referred by whom? Why they change a term?]{referred to as Monte Carlo (MC) Dropout}~\cite{gal2016dropout}.   

% Ensembling
Apart from Bayesian methods, ensembling-based approaches can be applied to uncertainty quantification as well~\cite{lakshminarayanan2017simple,wilson2020bayesian}.
However, vanilla ensembling methods is time and memory consuming because it is required to train and store multiple models.
To address this issue, Fast Geometric Ensembling (FGE)~\cite{garipov2018loss} and Stochastic Weight Averaging (SWA)~\cite{izmailov2018averaging} are proposed, which use varying learning rates during training to find different local minimums. 
In addition, model calibration methods such as Temperature Scaling~\cite{guo2017calibration} are also used to estimate prediction uncertainty.

Although uncertainty quantification has been quite popular in many deep learning domains, such as Computer Vision \cite{kendall2017uncertainties}, Medical Imaging \cite{abdar2021review} and Reinforcement Learning \cite{gawlikowski2021survey}, it is less explored in traffic prediction.
Wu et al. \cite{wu2021quantifying} analyzed different Bayesian and frequentist uncertainty quantification approaches for spatio-temporal forecasting. 
They figured that Bayesian methods were more robust in point prediction whilst frequentist methods provided better coverage over ground truth variations.
%This conclusion \task[(06-08) We don't have any result yet.]{is aligned with our results} as well.

In this paper, we specifically study the uncertainty quantification problem for spatio-temporal traffic prediction. 
The proposed approach is based on the spatio-temporal architecture \cite{bai2020adaptive} and combines Monte Carlo dropout, Adaptive Weight Averaging re-training, and model calibration
%\task[(06-08) You used ``a'' here. Then which one is it?]{a spatio-temporal model} and \task[(06-08) ``combine xxx, yyy, zzz, to what or for what?''] 
to provide both point prediction and uncertainty estimation surpassing current state-of-the-arts.

%%==================
\section{Problem Statement}
\label{sec:problem_statement}

%\todo[inline]{Need to be consistent. If deal with flow, title needs change and dataset needs to be flow. If not, all ``flow'' should be carefully re-examined.}
%\weizhu{The datasets used in the experiments are traffic flow, but the proposed method can be applied to other types of traffic data as well.}

Traffic data can be regarded as multivariate time series.
%Traffic \task[(06-08) ]{flow} data can be regarded as multivariate time series. 
Let $x_{t} \in \mathbb{R}^{N}$  be the values 
%\task[(06-08) ]{traffic flow values} 
of all the sensors in a road network at time $t$, and $X_{<t} = \{ x_{t-T_h+1}, x_{t-T_h+2}, \dots, x_{t}\} \in \mathbb{R}^{N \times T_h}$ be the corresponding historic input sequence with $T_h$ steps.
%where $T$ denotes the %\task[(06-08) We don't say like this.]{historic length}. 
Similarly, $\hat{X}_{>t} = \{ x_{t+1}, x_{t+2}, \dots, x_{t+\tau}\} \in \mathbb{R}^{N \times \tau}$ represents the prediction sequence, where $\tau$ denotes the prediction horizon.  
\begin{figure}[htbp]
\centering
\includegraphics[width=0.34\textwidth]{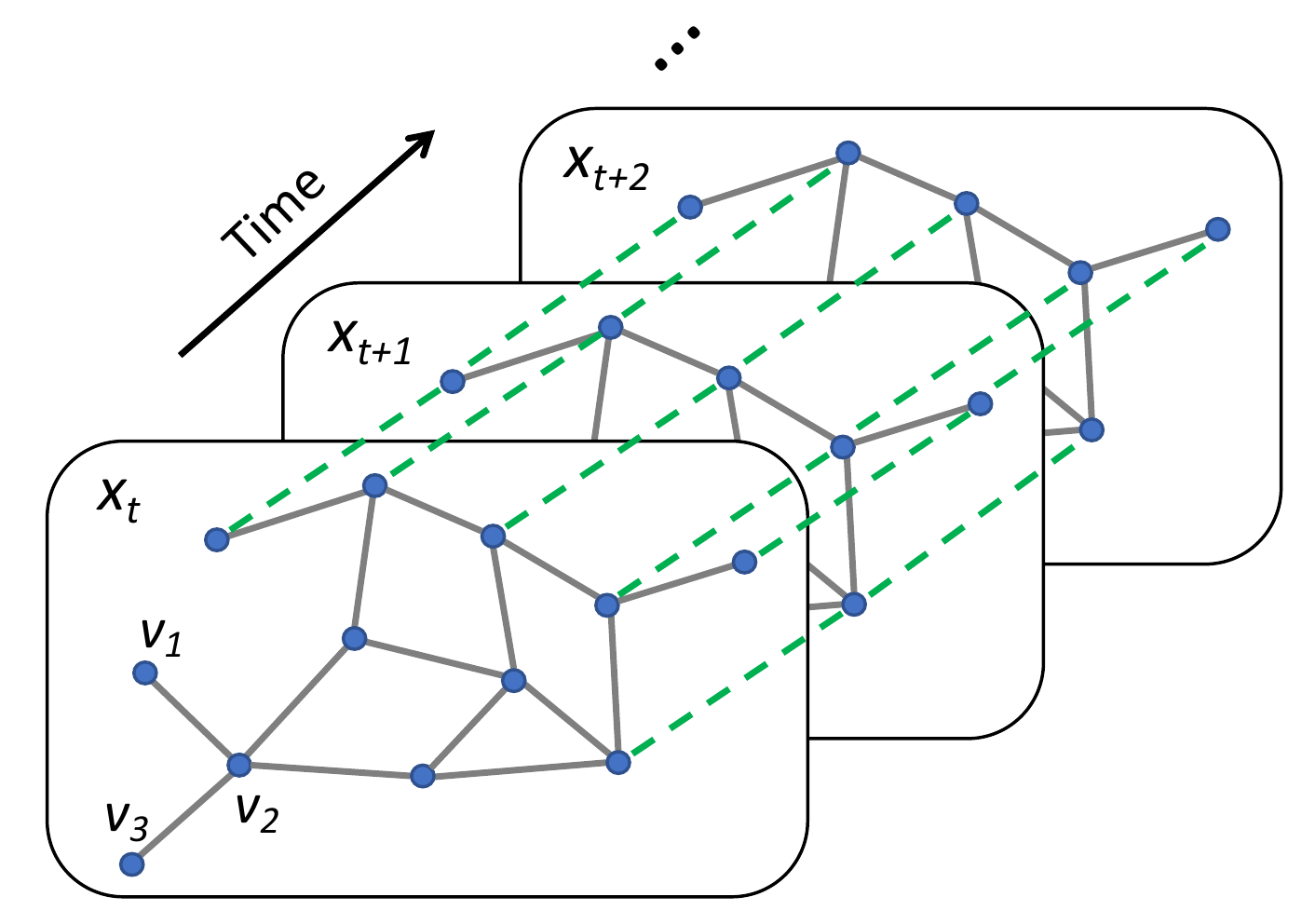}
\caption{Spatio-temporal dependency modelling for traffic data, where grey lines and green dash lines represent spatial and temporal dependency, respectively.}
\label{fig:problem}
\end{figure}
Fig. \ref{fig:problem} describes the spatio-temporal correlation modelling problem in traffic forecasting.
%\task[(06-08) Don't use ``our''. What's the difference between our task and the reviewer's task?]{our traffic flow forecasting task}.
%
%\task[(06-08) ``In particular'' means we have say one thing, let's say it in more details. It's not the case here.]{In particular}, 
Instead of treating the forecasting as deterministic, we aim to compute a conditional distribution to predict the traffic flow as well as the prediction uncertainty $\hat{X}_{>t}\sim P(\hat{X}_{>t}|X_{<t})$, which can improve the accuracy of the prediction, enhance the generalization ability of the model, and provide uncertainty estimation as well.

%The historic length is $1$ hour and prediction length is also $1$ hour. 

Note that although multi-modality and seasonality in traffic data do exist \cite{guo2019attention},  
%\task[(06-08) Reference]{}, their influence are more prominent when the time span is relatively large, e.g., $24$ hours. 
for simplicity, we do not consider multi modality or seasonality, and consequently treat the predictive distribution of each node at each time point as a conditional uni-variate Gaussian distribution. A similar assumption can also be found in \cite{zhu2017deep}. 
%\task[(06-08) Try to find other literature that made the same assumption.]{}
To this end, $P(\hat{X}_{>t}|X_{<t})$ can be represented by a set of predictive mean-variance pairs. As a result, the problem is cast as follows: 
%\task[(06-08) Yours?]{our problem} is formulated as follows:
\begin{align}
\label{eq:problem}
%\hat{X}_{>t}\sim \mathcal{N}(\hat{\mu} (X_{<t}),\hat{\sigma}(X_{<t})),
\theta = \underset{\theta}{\mathrm{argmax}} ~ \sum^N_{i=1} \log \mathcal{N}(\hat{X}^i_{>t};\hat{\mu}_\theta (X^i_{<t}),\hat{\sigma}_\theta(X^i_{<t})^2),
\end{align}
where $\theta$ is the model parameters, $N$ is the number of total training data points, $\mathcal{N}$ denotes the Gaussian likelihood, $\hat{\mu}(X_{<t})$ and $\hat{\sigma}(X_{<t})^2$ represents the estimated mean and variance, respectively.
%\task[(06-08) Equation (1) is not a prediction problem. It just describes a typical random variable. There is also nothing to optimize/learn in the equation.]{}

%%==================
\section{Deep Spatio-Temporal Uncertainty Quantification}
\begin{figure}[h]
\centering
\includegraphics[width=0.5\textwidth]{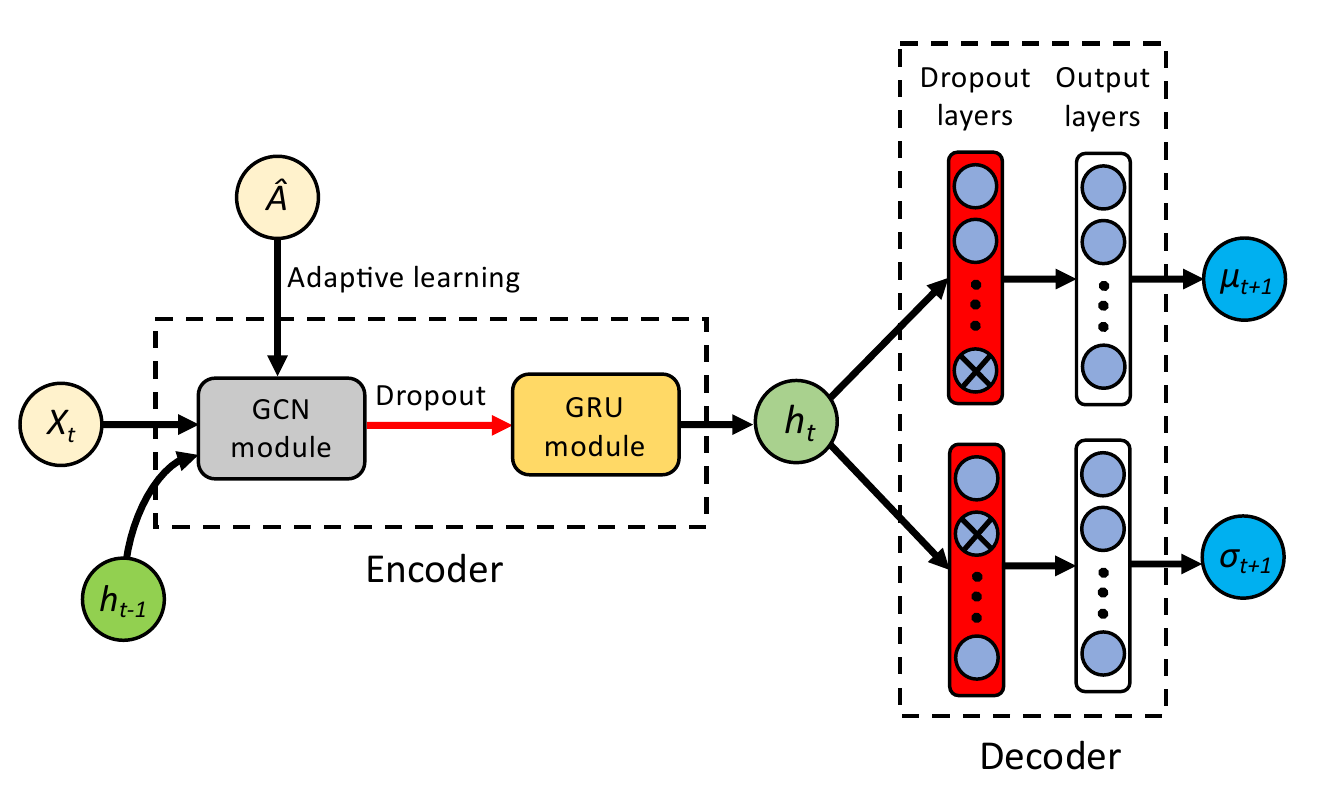}
\caption{Architecture of DeepSTUQ.}
\label{fig:architecture}
\end{figure}

%\section{\task[(06-14) Try to come up with a fascinating name and put it here and everywhere.]{Deep Spatio-Temporal Uncertainty Quantification}}
\label{sec:method}
We first briefly give an overview of the proposed Deep Spatio-Temporal Uncertainty Quantification (DeepSTUQ).  
The DeepSTUQ model architecture is illustrated in Fig. \ref{fig:architecture}, which follows the principle of \cite{bai2020adaptive}.
%\task[(06-14) A natural question: how your approach differs from the reference? May need to change, e.g., follows the design principle.]{which is based on \cite{bai2020adaptive}}.
This architecture includes an encoder and a decoder sub-neural network.
The encoder is composed of a GCN and a GRU module to capture both the spatial and temporal dependencies, respectively. 
To estimate the aleatoric uncertainty, the decoder employs two independent convolutional layers computing means and variances, respectively.  
Moreover, dropout operations are deployed in both sub-networks to estimate the epistemic uncertainty. 

% encoder-decoder
%\torevise{
%In the model, we let the input sequence and the previous hidden state pass through the GCN module with graph adaptive learning, to extract spatial features. 
%Then part of the spatial features is dropped binarily to generate uncertainty.  
%Afterward, we use a GRU module to capture the temporal dependency. 
%To generate more epistemic uncertainty, we let the obtained spatio-temporal features pass through a convolutional layer with dropout.
%Finally, we employ two independent convolutional layers compute the distribution of the final prediction by computing the corresponding means and variances.   
%}

In terms of model training, conventional training procedures are only capable of providing uni-modal solutions, 
%\task[(07-05) Reference or discussions to convince reader.]{which lacks diversity for quantifying uncertainty}.
which lacks diversity for quantifying uncertainty \cite{gawlikowski2021survey}.
To address this issue, a three-stage training method is proposed, which can be briefed as follows.
\begin{itemize}
    \item \textit{Stage 1}: Pre-train the base spatial-temporal model with dropout on the training dataset to perform variational learning;
    \item \textit{Stage 2}: Re-train the pre-trained model on the training dataset to proceed ensemble learning;   
    \item \textit{Stage 3}: Calibrate the re-trained model on the validation dataset to further improve the aleatoric variance estimation.  
\end{itemize}
%\task[(06-14) May need to explain the difficulty using typical training methods, so you need to propose a new one.]{we propose} a three-stage training method, which can be briefed as follows.
In the following sections, we will introduce DeepSTUQ in detail.

%%==================
\subsection{Spatial Dependency}
\label{sec:spatial}

\subsubsection{Graph Convolution}
A typical road network consists of a number of road segments. 
The spatial relationships within a road network with $N_R$ road segments can be described through a graph $\mathcal{G(V, E)}$, where the nodes $\mathcal{V} = \{v_1, v_2, v_3, \cdots, v_{\lvert{}\mathcal{V}\rvert{}} \}$
%\task[(06-14) Can we change $N_R$ to $\lvert{}V\rvert{}$? So we do not need to introduce a new symbol $N_R$.]{} 
denote the sensors and the edges $\mathcal{E}$ denotes the road segment. 
$A \in \mathbb{R}^{\lvert{}\mathcal{V}\rvert{} \times \lvert{}\mathcal{V}\rvert{}}$ is the corresponding adjacency matrix.
Subsequently, GCN \cite{kipf2016semi} is utilized to model the spatial relationships of the traffic data.
The output of the $l\text{-th}$ GCN layer, $Z^{(l+1)}$, can be computed by 
\begin{align}
\label{eq:gnn}
Z^{(l+1)} & = f(Z^{(l)}, A),
\end{align}
where $Z^{(l)}$ is the input.
%
%\todo[inline]{add variable dimension}
%
%\task[(06-14) Break down into small sentences. No more than two subordinate clauses in one sentence.]{ More specifically, to implement \ref{eq:gnn} via neural networks, following the convention of GCNs \cite{kipf2016semi}, we use an identity matrix $I$ to sum up the neighboring nodes of a node as well as the node itself, and a degree matrix $D$ to avoid changing the scale of feature vectors when multiplying with $A$.} 
More specifically, GCN first uses a degree matrix $D$ to avoid changing the scale of feature vectors by multiplying it with $A$. 
Afterwards, an identity matrix $I$ is used to sum up the neighboring nodes of a node as well as the node itself.
As a result, the propagation rule of GCN is described as follows:  
\begin{align}
\label{eq:gcn}
Z^{(l+1)} = S \big ((I + D^{-\frac{1}{2}} A D^{-\frac{1}{2}})Z^{(l)}W^{(l)}+b^{(l)}\big),
\end{align}
where $W^{(l)}$ is the weight matrix, $b^{(l)}$ is the bias, and $S$ is an activation function, e.g., sigmoid function.

%%================
\subsubsection{Graph Structure Learning}
In many real-world cases, %\task[(06-14) Very rare. More commonly we know the connectivity, but connectivity is not correlation. This is a quite weak point. Please change over all places.]{we do not have the knowledge} of the graph structures. 
we do not have the real spatial correlation knowledge of the multivariate traffic data.
In such cases, the graph structure needs to be learned from data.
To this end, the adaptive learning approach proposed in \cite{bai2020adaptive} is adopted to directly generate $\hat{A} = D^{-\frac{1}{2}}A D^{-\frac{1}{2}}$, which is easier than generating the adjacency matrix during the training process. 
Particularly, $\hat{A}$ is developed by 
\begin{align}
\label{adaptive_gcn}
\hat{A} = \operatorname{softmax} \big(\operatorname{ReLU}(EE^T)\big),
\end{align}
where $E \in \mathbb{R}^{\lvert{}\mathcal{V}\rvert{} \times d}$ (the embedding dimension $d \ll \lvert{}\mathcal{V}\rvert{}$) is a learnable matrix representing the embedding of the nodes and %\task[(06-14) Change all this to $\operatorname{softmanx}$]{$softmax$} 
$\operatorname{softmax}$ function is to normalize the learned matrix. 
To facilitate the graph learning process, the Node Adaptive Parameter Learning (NAPL) module \cite{bai2020adaptive} is also utilized to reduce the computational cost. As a result, Equation \eqref{eq:gcn} becomes
\begin{align}
\label{eq:napl_gcn}
Z^{(l+1)} = S \big ((I + \hat{A})Z^{(l)}EW_g^{(l)}+Eb_g^{(l)}\big).
\end{align}

%%==================
\subsection{Temporal Dependency}
\label{sec:temporal}
%\subsection{Gated Recurrent Unit}
Apart from the spatial dependency, the temporal dependency of traffic data also needs to be captured.
To this end, the aforementioned graph convolutional operations and adaptive graph learning 
%\task[(06-14) Reads strange. Please re-phrase.]{into a model structure based on} 
module are integrated into a Gated Recurrent Unit (GRU) \cite{cho2014learning}.
Subsequently, the obtained spatio-temporal model can be formulated as follows:    
\begin{subequations}
\label{eq:gru}   
\begin{align}
%& \tilde{A} = softmax(ReLU(E \cdot E^T))\\
& z_t = S \big ((I+\hat{A})[x_{t},h_{t-1}]EW_z+Eb_z\big),\\
& r_t = S \big ((I+\hat{A})[x_{t},h_{t-1}]EW_r+Eb_r\big),\\
& c_t = \tanh \big((I+\hat{A})[x_{t},r_t\odot h_{t-1}]EW_c + Eb_c\big),\\
& h_t = z_t \odot h_{t-1} + (1-z_t) \odot c_t,
\end{align}
\end{subequations}
where $z$ stands for the update gate, $r$ stands for the reset gate, $h$ denotes the hidden state, $\operatorname{[\cdot]}$ denotes the concatenation operation, $c$ denotes the memory cell, $W$ and $b$ represent the weights and bias, respectively.

Finally, the model introduced in Equation \eqref{eq:gru} serves as the spatio-temporal architecture in DeepSTUQ.
Note that though the above base model is employed in this work, DeepSTUQ has the potential to be applied to other spatial-temporal structures as well. 
In the following sections, we explain how to leverage this base model to forecast traffic and quantify the corresponding forecasting uncertainty.

%%==================
\subsection{Uncertainty Quantification}
Generally, uncertainty can be classified into two types, i.e., epistemic and aleatoric. The former represents model uncertainty, while the latter represents data uncertainty. 
If variance 
%\fix[Try to avoid using ``we do something''. Instead, use subjective voice. Please re-phrase the whole writing.]{}{is used }
is used to render uncertainty, the total uncertainty can be decomposed and approximated as follows:
% \begin{subequations}
% \begin{align}
% \label{eq:uncertainty}
% \text{Aleatoric uncertainty}  & \approx \mathbb{E}_{\theta \sim p(\theta)}[\sigma^2_\theta],  \\
% \text{Epistemic uncertainty}  & \approx  \mathbb{V}_{\theta \sim p(\theta)}[\mu_\theta], 
% \end{align}
% \end{subequations}
\begin{align}
\label{eq:uncertainty}
\sigma_\text{Total}^2  \approx \underbrace{\mathbb{E}_{\theta \sim p(\theta)}[\sigma^2_\theta] }_{\text{Aleatoric uncertainty}} + 
\underbrace{\mathbb{V}_{\theta \sim p(\theta)}[\mu_\theta]}_{\text{Epistemic uncertainty}}, 
\end{align}
where $p(\theta)$ stands for a probability distribution over the model parameters $\theta$, $\sigma^2_\theta$ and $\mu_\theta$ refer to predicted variance and mean, respectively.
%\thought{This equation is more suitable for ensembling approach.}

%%==================
\subsubsection{Aleatoric Uncertainty}
Aleatoric uncertainty is caused by the intrinsic randomness of data, which is irreducible but learnable \cite{abdar2021review}.
%\task[(06-14) Reads strange.]{is data-dependent or system-dependent},
%\task[(06-14) Provide a reference]{}. 
%In this work, we focus on data-dependent aleatoric uncertainty, namely, heterogeneous aleatoric uncertainty. 
Based on Equation \eqref{eq:uncertainty}, 
%\task[(07-05) Still not strong. Here implies regression=Gaussian. Maybe add a footnote to explain this common practice.]
we assume that the lower and upper bounds of the forecasting are symmetric due to the regressive nature of the prediction. Subsequently,
%\task[(06-14) Why Gaussian? Any support? \\ Weizhu: Gaussianity is a commonly used assumption, it works well in practice. For further study, this can be replaced by some distribution-free methods.]{assume} 
the distribution of a sensor's value, e.g., traffic flow, at each time point can be modeled by a Gaussian distribution with predicted mean $\mu(x)$ and variance $\sigma(x)^2$.
However, directly maximizing the predictive Gaussian likelihood is numerically unstable. Instead we choose to maximize the following log-likelihood:   
\begin{align}
\label{eq:heter_uncertainty}
\log &\; p(y|\mu(x), \sigma(x)) \nonumber\\
%=&\, \log (\frac{1}{\sigma(x) \sqrt{2 \pi}}e^{-\frac{1}{2}(\frac{y-\mu(x) }{\sigma(x)})^2}) \nonumber\\
%=&\,\log (\frac{1}{\sigma(x,A) \sqrt{2 \pi}}) + \log (e^{-\frac{1}{2}(\frac{x-\mu(x)}{\sigma(x)})^2}) \nonumber\\
=&\, -\frac{1}{2}\log (\sigma(x)^2) - \frac{1}{2} \log (2 \pi) -\frac{(y-  \mu(x))^2}{2\sigma(x)^2}, 
\end{align}
where $\log (\sigma(x)^2)$ and $\mu(x)$ are obtained directly via two independent neural networks.  

In practice, to accelerate the training process and ensure convergence, we devise the following weighted loss by adding an L1 loss as the regularization term based on Equation \eqref{eq:heter_uncertainty}:
\begin{align}
\label{eq:heter_loss}
\mathcal{L_\text{Aleatoric}} = \frac{1}{N} \sum^N_{i=1} \lambda &\,\big \{\log (\sigma(x_i)^2)+\frac{(y_i - \mu(x_i))^2}{\sigma(x_i)^2} \big \}  \nonumber\\
&\,+ (1-\lambda) |y_i-\mu(x_i)|,
\end{align}
where $\lambda$ is the relative weight with $0 < \lambda \leq 1$.

%%==================
\subsubsection{Epistemic Uncertainty}
\begin{figure}[htbp]
\centering
\includegraphics[width=0.42\textwidth]{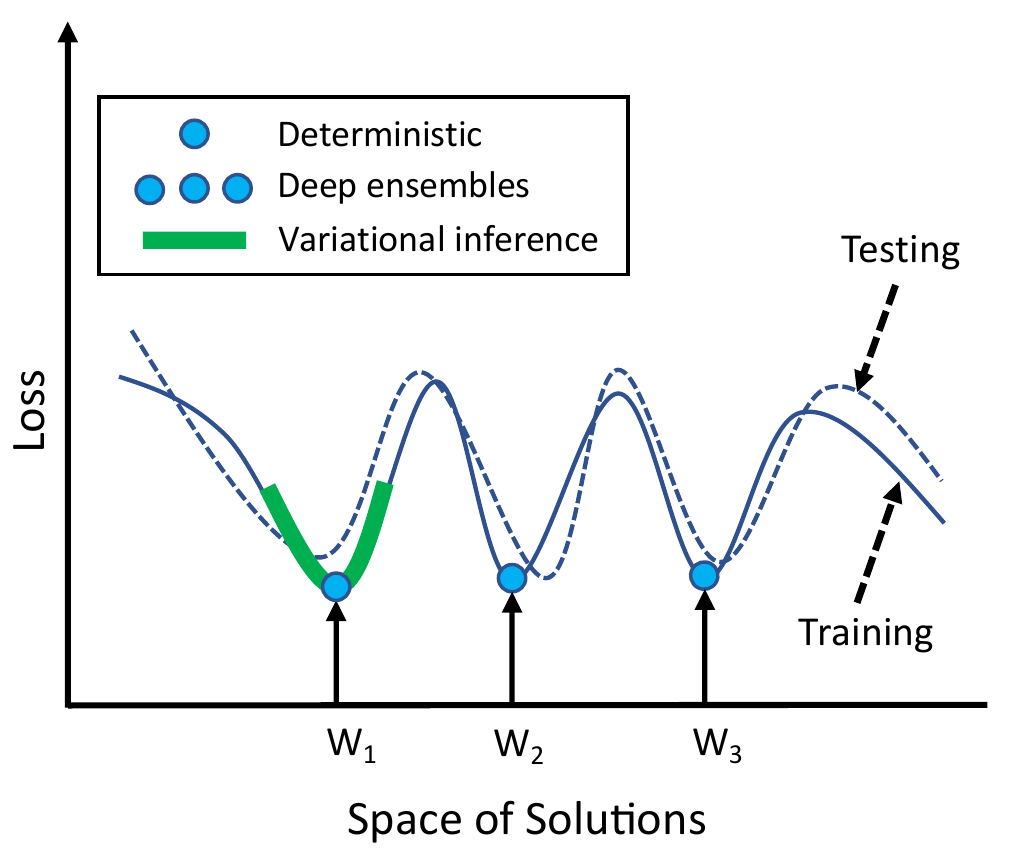}
\caption{Performance demonstration of deterministic model, deep ensembles, and variational inference in solution space.}
\label{fig:solution_space}
\end{figure}

Epistemic uncertainty represents model uncertainty, which arises from that lack of data or model mis-specification. 
Fortunately, as opposed to aleatoric uncertainty, epistemic uncertainty can be reduced by estimation.% by Bayesian approximation and/or ensembling. \task[(06-14) One sentence introducing Fig. 3 and how to read it.]{}
There are two general classes of approaches to do so: Bayesian variational inference and deep ensembling.
However, they both have their pros and cons.
Fig. \ref{fig:solution_space} illustrates the relationships between different solutions and corresponding model performance. 
The solid and dashed lines represent the model performance during training and testing processes, respectively.
The green line and blue dots represent the performance that can be obtained by variational inference and deterministic model, respectively.
As it can be seen from the figure, deep ensembling can find a set of different deterministic model parameters (local minimums), e.g., $W_1$, $W_2$, and $W_3$, which may have equally good performance in the solution space \cite{fort2019deep}. 
On the other hand, variational inference can find a set of sub-optimal solutions near one local minimum in the loss space. 
However, it may fail to find other local minimums, which potentially leads to modal collapse.
Therefore, a better way is to explore as many as local minimums as well as their corresponding nearby solutions.   
To this end, %deep ensembling and variational inference are combined to estimate epistemic uncertainty. 
we propose to combine deep ensembling and variational inference to estimate epistemic uncertainty.

%As it can be seen from the figure, deep ensembling can find a set of different deterministic model parameters (local minimums), e.g., $W_1$, $W_2$, and $W_3$, which may have equally good performance in the solution space \cite{fort2019deep}. On the other hand, \task[(06-14) You need to first introduce that there are two main streams. Then introduce their merits and defects.]{variational inference} can find a set of sub-optimal solutions near one local minimum in the solution space. However, it may fail to find other local minimums, which potentially leads to modal collapse. 
%\task[(06-14) And then? What is the conclusion drawn from the above analysis?]{}

%\task[(06-14) Why here?]{We adopt two types of approaches to obtain model uncertainty, namely, variational inference and deep ensembling. }
%As illustrated in Fig. \ref{fig:solution_space}, 
%to combine the advantages of these two methods, we combine deep ensembling with variational inference in our approach to obtain better estimation of epistemic uncertainty.  
%\todo[inline]{Re-write the above paragraph. It seems something is wrong.}

%%==================
\textbf{Variational Inference}.
Let $D=\{X, Y\}$ be the training dataset. From a Bayesian perspective, we assume each weight parameter of the neural network $w$ obeys a probabilistic distribution to represent model uncertainty, e.g., Gaussian distribution. 
However, in practice, the true posterior of the the neural network weights $p(w|D)$ is intractable. Therefore, a variational distribution $q(w)$ is used to approximate $p(w|D)$. 
Accordingly, %\task[(06-14) Try also avoid ``our'' throughout the whole paper.]{our} 
the optimization goal is to minimize the following Kullback-Leibler (KL) divergence: 
\begin{align}
\label{eq:elbo}
& D_\text{KL}(q(w)||p(w|D)) \nonumber \\ 
& = \int q(w) \log \frac{q(w)}{p(w)p(D|w)} dw, \nonumber \\ 
& = D_\text{KL}(q(w)||p(w)) - \mathbb{E}_{w \sim q(w)} [\log p(D|w)],
\end{align}
where $p(w)$ is the prior and $\log p(D|w)$ is the predictive log-likelihood.

% From a Bayesian perspective, to quantify epistemic uncertainty, we assume each weigh of the neural network, $w$, is with respect to a probabilistic distribution, $p(w)$,  i.g., Gaussian distribution. Let $D=\{X, Y\}$ be the training dataset, $\hat{x}$ be a testing data point, then the joint posterior distribution can be represented as follows:
% \begin{align}
% \label{eq:bnn}
% p(\hat{y}|\hat{x},D) = \int p(\hat{y}|\hat{x},w) p(w|D) dw, 
% \end{align}
% where $\hat{y}$ is the predictive mean, $p(\hat{y}|\hat{x},w)$ is the predictive likelihood, and $p(w|D)$ is the true posterior of the weights. 
% In practice, $p(w|D)$ is usually intractable, so we need to utilize a variational distribution $q(w)$ to approximate it. 

%\task[(06-14) Re-phrase. Too many subordinate clauses.]
%for the sake of simplicity and flexibility, 
%MC Dropout \cite{gal2016dropout} is a simple and flexible Bayesian approximation approach
To solve Equation \eqref{eq:elbo}, MC dropout \cite{gal2016dropout} is adopted as it performs Bayesian approximation in a simple and flexible manner. 
%by using dropout \cite{srivastava2014dropout} to formulate the variational distribution $q(w)$. 
The variational distribution $q(w)$ formulated in MC dropout can be described as follows.
Let $W_i$ be a matrix of shape $K_{j} \times K_{j-1}$ for layer $i$, we have
\begin{subequations} 
\begin{align}
\label{eq:dropout}
q(W_i) & = M_i \cdot (\operatorname{diag}[z_{i,j}]^{K_i}_{j=1}), \\
z_{i,j} & \sim \operatorname{Bernoulli}(p_i)
\end{align}
\end{subequations}
%\task[(06-14) diag and Bernoulli should be in operatorname]{}
where $W_i$ denotes the masked weight matrices, $p_i$ is dropout rate used in both the training and testing processes (as opposed to standard dropout), $M_i$ is the parameters of the neural network in the $i$-th layer , and $z_{i,j}$ is a binary variable indicating whether unit $j$ at layer $i-1$ (as the input of layer $i$) is dropped.
%\task[(06-14) Why $i-1$ instead of $i$?]{$i-1$} is dropped. 
As a result, minimizing Equation \eqref{eq:elbo} is equivalent to minimizing the following loss function:  
\begin{align}
\label{eq:droput_loss}
\mathcal{L_\text{Dropout}} & = \mathbb{E}_{w \sim q(w)} E[Y,f_W(X)] - D_\text{KL}(q(w)||p(w)), \nonumber \\ 
& \approx \frac{1}{N} \sum_{i=1}^N E(y_i,f(x_i, w_i)) + \frac{\lambda_W}{2p_i}||w_i||^2,
\end{align}
where $E$ is the loss function, e.g., Root Mean Squared Error (RMSE) or Mean Absolute Error (MAE), $\lambda_W$ is the weight decay, and
$\frac{\lambda_W}{2p}||w||^2$ can be computed through applying the L2 regularization during the training process. 

In terms of implementation, dropout operations are deployed at two places within the spatial-temporal model: the graph convolutional layers in the encoder and the dropout convolutional layers in the decoder. 
Therefore, Equation \eqref{eq:napl_gcn} becomes
%\begin{subequations}
\begin{align}
\label{eq:adj_dropout}
%X^l = \sigma(Z \odot (softmax(ReLU(E \cdot E^T)) X^{l-1}) W), 
Z^{(l+1)} = \operatorname{sigmoid} \Big ( M \odot \big ((I + \hat{A})Z^{(l)}EW_g^{(l)}+Eb_g^{(l)}\big)\Big).
% & z_t = \sigma \Big ( M_z \odot\big ((I+\hat{A})[X_{:,t},h_{t-1}]EW_z+Eb_z\big)\Big)\\
% & r_t = \sigma \Big ( M_r \odot\big (I+\hat{A})[X_{:,t},h_{t-1}]EW_r+Eb_r\big)\Big)\\
% & c_t = \tanh \Big ( M_c \odot\big (I+\hat{A})[X_{:,t},r_t\odot h_{t-1}])EW_c + Eb_c\big)\Big) \\
% & h_t = M_c \odot \big ( z_t \odot h_{t-1} + (1-z_t) \odot c_t \big),
\end{align}
%\task[(06-14) ReLU in operatorname]{}
%\end{subequations}
Note that the dropout rate here should be small when the adjacency matrix dimension is small, and vice versa.
 
%%==================
\textbf{Combined Uncertainty}. Finally, 
%\task[(06-14) Change Eq. ref(xxx) to eqref{xxx}. For all places in the text.]
Equations \eqref{eq:droput_loss} and \eqref{eq:heter_loss} are combined to estimate both aleatoric and epistemic uncertainty jointly. 
The combined loss function is formulated by   
\begin{align}
\label{eq:combined_loss}
\mathcal{L_\text{Combined}} = \frac{1}{N} & \sum_{i=1}^N  \lambda \,\big \{\log (\sigma(x_i)^2)+\frac{(y_i - \mu(x_i))^2}{\sigma(x_i)^2} \big \}  \nonumber\\
&\,+ (1-\lambda) |y_i-\mu(x_i)| + \frac{\lambda_W}{2p}||w||^2. 
\end{align}
Equation \eqref{eq:combined_loss} is utilized to pre-train the spatio-temporal model in DeepSTUQ . 

%%==================
\textbf{Deep Ensembling}.   
In contrast to variational inference, deep ensembling aims to find a set of different local minimums and averages the output of each trained model as the final prediction. 
Deep ensembling is shown to be quite effective in practice, yet it is computationally expensive as multiple models are trained \cite{lakshminarayanan2017simple}. 
FGE \cite{garipov2018loss} tackles this issue by using cycling learning rate to produce a set of different trained models in one learning process. 
However, FGE still needs to store multiple models for inference, which may result in high memory cost. 
To address this issue, SWA \cite{izmailov2018averaging} adjusts the learning rate and averages the weights during the learning process to generate only one trained model to approximate FGE. 
In SWA, the model parameters are updated by 
\begin{align}
\label{eq:swa_update}
w_\text{SWA} = \frac{w_\text{SWA}\cdot n_\text{models}+w}{n_\text{models}+1},  
\end{align}
where $w_\text{SWA}$ is the parameters of the SWA model and $n_\text{models}$ is the number of averaged models during training. 

\begin{figure}[htbp]
\centering
\includegraphics[width=0.47\textwidth]{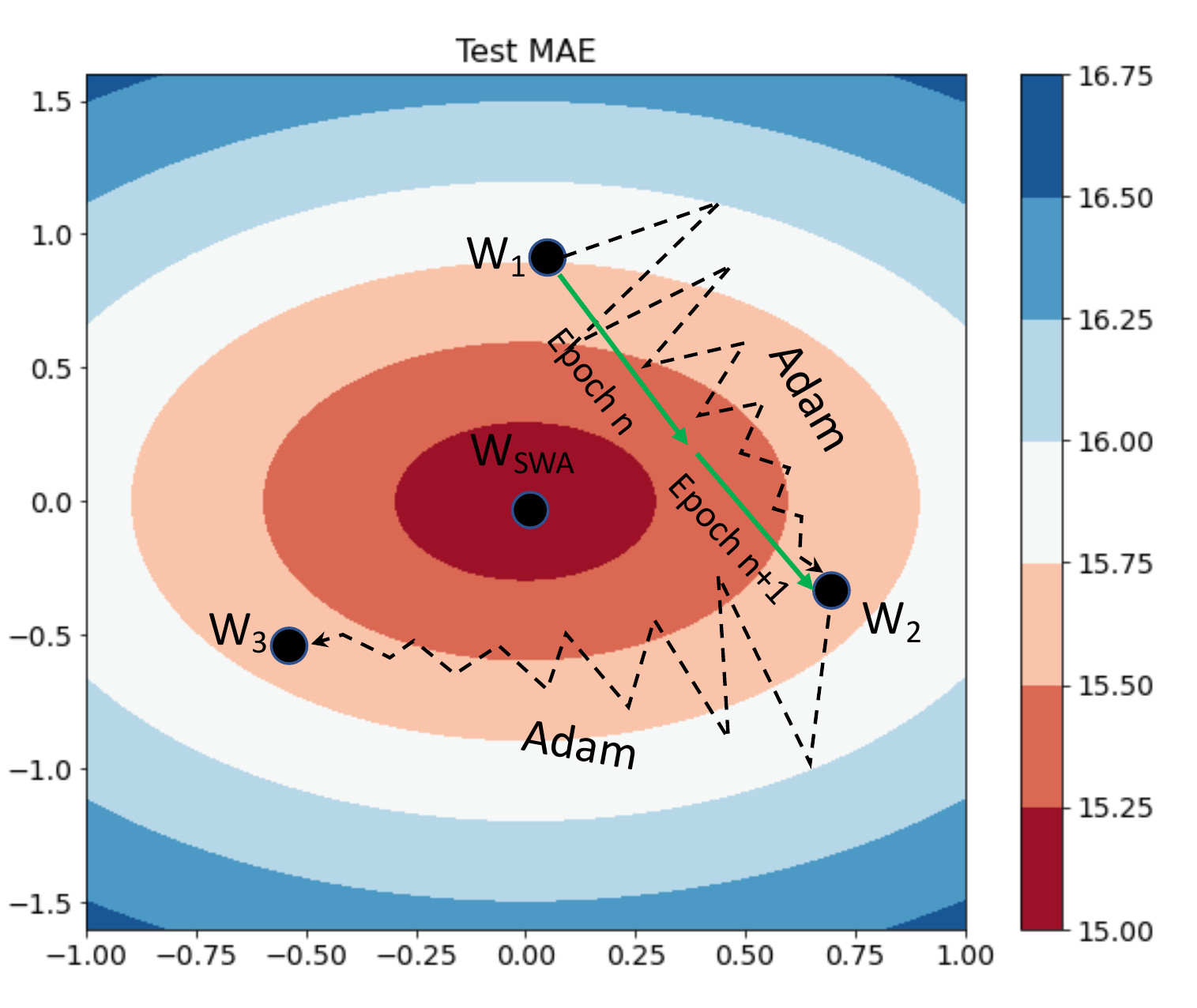}
\caption{Demonstration of relationship between test MAEs and model weights during AWA re-training.}
\label{fig:swa}
\end{figure}

%\torevise{
Inspired by SWA, we devise a re-training method called Adaptive Weight Averaging (AWA) to approximate deep ensembling. 
%Specifically, \task[(06-14) Strange why saying this before introducing your method.]{we find that in practice using Adam as the optimizer works more effectively than using Stochastic Gradient Decent (SGD) which is adopted in the original SWA method \cite{izmailov2018averaging}.}
As depicted in Fig. \ref{fig:swa}, we vary the learning rate during the re-training process to find different local minimums, and average those local minimums in the final stage to attain better solutions. 
%The proposed \task[(06-14) Give it a name without SWA. The current presentation implies no novelty.]{SWA-style} learning approach includes two steps. 
The proposed AWA re-training approach includes two steps.
Let the re-training learning rate be $lr$, the maximum learning rate be $lr_1$, the minimum learning rate be $lr_2$, $n_\text{iteration}$ be the total iteration number within each epoch/total batch number, then the learning rate at $n_i$ iteration changes according to the following rules.
%\task[(06-14) Into small sentences.]{At the first step, to enable the trained model to escape from the current local minimum, we let the learning rate of the optimizer, starts from $lr_1$ and decrease gradually to $lr_2$ using a cosine learning rate scheduler at epoch $n$, which can be described by}
The first step is to enable the trained model to escape from the current local minimum.
To this end, the learning rate of the optimizer decreases from $lr_1$ to $lr_2$ via a cosine learning rate scheduler at epoch $n$. The scheduler is described by
\begin{align}
\label{eq:learning_rate1}
lr = lr_2 + \frac{1}{2}(lr_1-lr_2) \big(1+\cos(\frac{n_\text{iteration}}{n_i}\pi)\big).
\end{align}

\begin{figure}[htbp]
\centering
\includegraphics[width=0.48\textwidth]{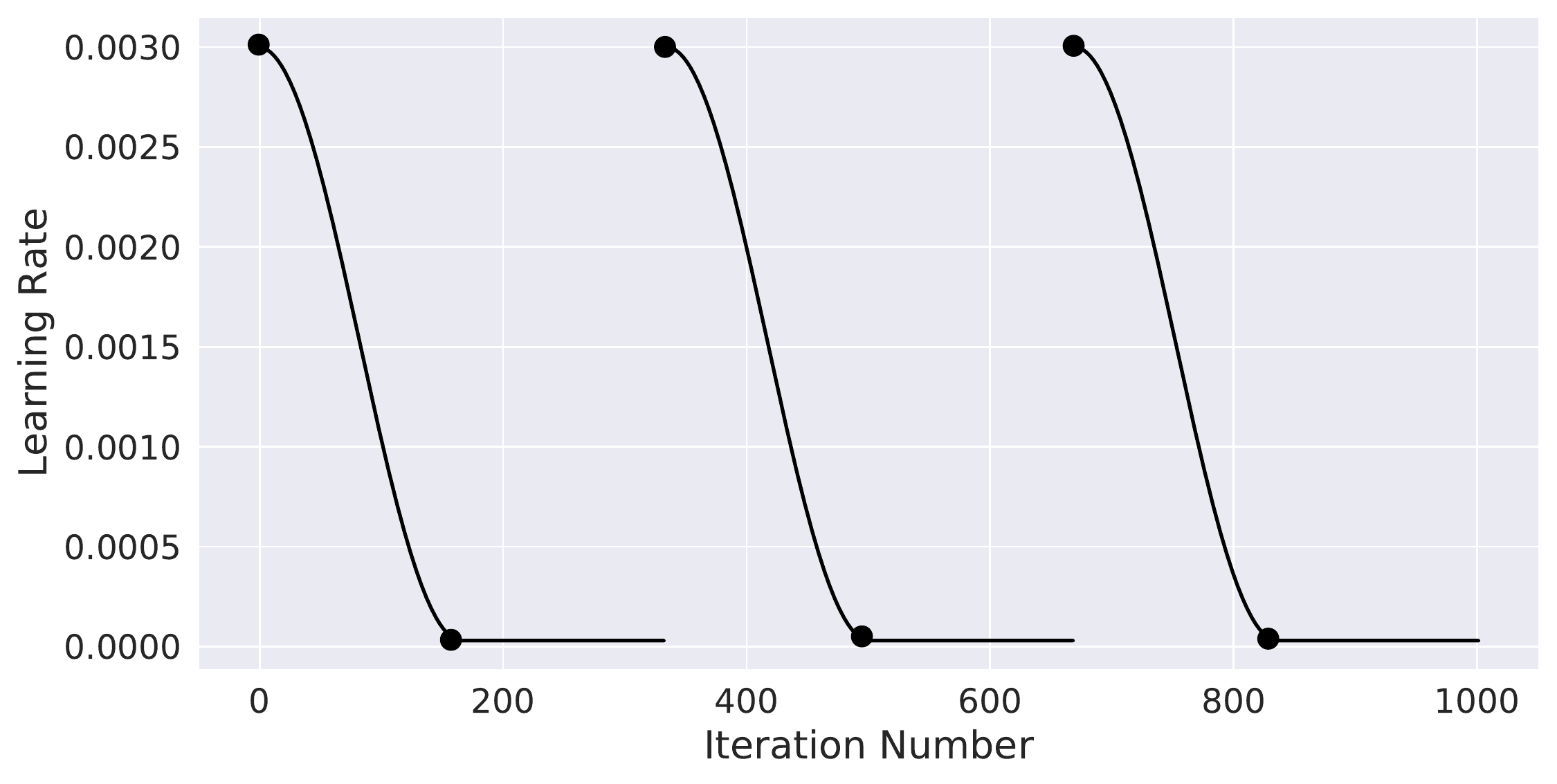}
\caption{\footnotesize Learning rate change during the AWA re-training, each black dot indicates the start of a new epoch.}
\label{fig:learn_rate}
\end{figure}

Following that, the model is fine tuned by using the constant learning rate $lr_2$ at epoch $n+1$, then at the end of the epoch the model parameters are averaged according to Equation \eqref{eq:swa_update} and perform batch normalization. 
%\begin{subequations}
% \begin{align}
% \label{eq:learning_rate2}
% lr = lr_2.  
% \end{align}
%\end{subequations}
Specifically, we find that in practice using Adam as the optimizer works more effectively than using Stochastic Gradient Decent (SGD) which is adopted in the original SWA method.
The learning rate change during the AWA re-training is illustrated in Fig. \ref{fig:learn_rate}.
The whole re-training process is summarized in Algorithm \ref{alg:swa_training}.

\begin{algorithm}[htbp]
\caption{AWA Re-training Method}
\label{alg:swa_training}
\textbf{Require}: training dataset \{$X$\}; pre-trained model parameters $w$; AWA model parameters $w_\text{AWA}$; learning rates: $lr_1$ and $lr_2$; total epoch $\text{epoch}_\text{AWA}$; total iteration/batch number $n_\text{iteration}$.
\begin{algorithmic}[1] 
\STATE{\textbf{while} $\text{epoch} < \text{epoch}_\text{AWA}$:}
   \STATE{\quad \textbf{while} $n <n_\text{iteration}$:}
        \STATE{\quad \quad  compute the loss function according to Equation \eqref{eq:combined_loss} and update $w$;}
        \STATE{\quad \quad \textbf{if} epoch $//$ 2 = 0:}
        \STATE{\quad \qquad  $lr$ decreases from $lr_1$ to $lr_2$ according to Equation \eqref{eq:learning_rate1};}
        \STATE{\quad \quad \textbf{else}: $lr = lr_2$;}
  \STATE{\quad \textbf{end while}}  

    \STATE{\quad \textbf{if} epoch $//$ 2 = 0 and epoch $\neq$ 0:}    
        \STATE{\quad \quad update $w_\text{AWA}$ according to Equation \eqref{eq:swa_update};}
        \STATE{\quad \quad perform batch normalization.}
\STATE{\textbf{end while}}          
\STATE{\textbf{Return} $w_\text{AWA}$}
\end{algorithmic}
\end{algorithm} 

At test time, we quantify the epistemic uncertainty by drawing multiple Monte Carlo samples from the learnt posterior distribution, then use the means and variances of the samples as the predictive mean and variances, respectively.

%%==================
\subsubsection{Model Calibration}
%\question{Purpose of model calibration?}
To prevent the uncertainty estimation of the trained models being overconfident on the training dataset, it is necessary to calibrate the trained model on the validation dataset with post-processing.
%To this end, we adapt the method proposed in \cite{guo2017calibration} \task[(06-14) ?]{to a regression setting to re-calibrate} the predicted aleatoric uncertainty obtained by \eqref{eq:heter_loss} to provide more accurate uncertainty estimation
%\weizhu{To this end, we leverage a TS approach to re-calibrate the predicted aleatoric uncertainty obtained by \eqref{eq:heter_loss}.}

%\task[(06-14) Count how many ``to'' you have in this sentence.]{}. 
%After the model is trained, %\task[(06-14) ]{we}
%it is re-calibrate on the validation dataset. %\task[(06-14) ]{We} 
To this end, a positive learnable variable $T$ is imposed on the learned variance.
Subsequently, the following log-likelihood similar to Equation \eqref{eq:heter_uncertainty} is maximized:
\begin{align}
\label{eq:ll_ts}
\log &\; p(y|\mu(x), \sigma(x)/T) \nonumber\\
=&\, \log (\frac{T}{\sigma(x) \sqrt{2 \pi}}e^{-\frac{1}{2}((\frac{T(y-\mu(x)) }{\sigma(x)})^2}) \nonumber\\
=&\,\log (\frac{T}{\sigma(x) \sqrt{2 \pi}}) + \log (e^{-\frac{1}{2}(\frac{T(y-\mu(x))}{\sigma(x)})^2}) \nonumber\\
=&\, -\frac{1}{2}\log (\frac{\sigma(x)^2}{T^2}) - \frac{1}{2} \log (2 \pi) -\frac{T^2(y- \mu(x))^2}{\sigma(x)^2} \nonumber\\
=&\, \frac{1}{2}\log (T^2)  -\frac{1}{2}\log (\sigma(x)^2) -\frac{T^2(y- \mu(x))^2}{2\sigma(x)^2} - \frac{1}{2} \log (2 \pi), 
\end{align}
where $T$ is the only learnable parameter. 
Accordingly, the calibration objective is 
\begin{align}
\label{eq:loss_cali}
T = \underset{T}{\mathrm{argmin}} \frac{1}{N} \sum_{i=1}^N - & \log (T^2)  + \frac{T^2(y_i- \mu(x_i))^2}{\sigma(x_i)^2},
\end{align}
where $\mu(x_i)$ and $\sigma(x_i)^2$ can be obtained via one deterministic forward pass or Monte Carlo estimation. Limited-memory Broyden-Fletcher-Goldfarb-Shanno algorithm (L-BFGS) is used as the optimizer to find the optimal value of $T$. 

\subsection{Proposed Unified Approach}
Finally, combining the spatio-temporal correlation modelling method, Monte Carlo dropout, AWA re-training, and model calibration, the proposed unified uncertainty quantification method can be summarized as follows. 

\begin{itemize}
    \item First, %\task[(06-14) ]{we} 
    the spatio-temporal model introduced in Section \ref{sec:spatial} and \ref{sec:temporal} is pre-trained using Equation~\eqref{eq:combined_loss} as the training loss function on the training dataset to estimate the aleatoric and epistemic uncertainty;
    \item Afterwards, the pre-trained model is re-trained via the AWA method on the training dataset to approximate deep ensembling;
    %Once the model is trained, \task[(06-14) ]{we} use the proposed \task[(06-14) ]{SWA-style approach} to re-train our model on the training dataset to approximate deep ensembling;
    \item Finally, %\task[(06-14) ]{we} 
    the predicted $\sigma^2$ obtained via the re-trained model on the validation dataset is calibrated according to Equation~ \eqref{eq:loss_cali}.
\end{itemize}

\begin{figure}[htbp]
\centering
\includegraphics[width=0.3\textwidth]{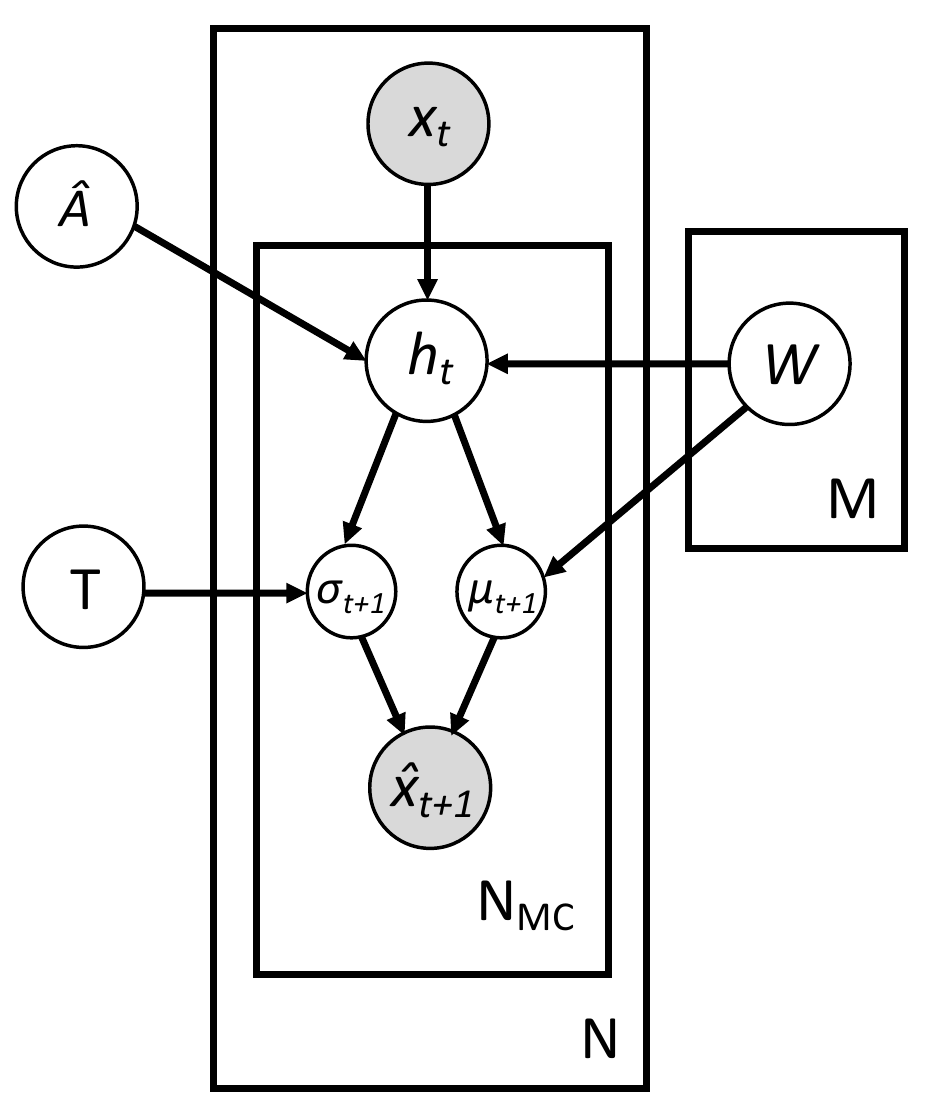}
\caption{Graphical model representation of DeepSTUQ, where shaded circles represent observable variables, arrows denote dependencies, variables within rectangles appear repeatedly, $N_{MC}$ is the number of Monte Carlo samples, and $M$ is the number of models for ensembling.}
\label{fig:plate_model}
\end{figure}

The graphical probabilistic model representation of DeepSTUQ is visualized in Fig. \ref{fig:plate_model}. 
The figure shows that $h_t$ is extracted from $x_t$ via a spatio-temporal structure with a learnable variable $\hat{A}$. 
The model weights are drawn repeatedly to estimate the epistemic uncertainty, which is implemented in an efficient manner by using MC dropout and AWA.   
The variance $\sigma(x_i)^2$ and mean $\mu(x_i)$ are obtained via $N_{MC}$ Monte Carlo samples. 
Finally, $\sigma(x_i)^2$ is calibrated through learning an auxiliary variable $T$.

At test time, according to Equation \eqref{eq:uncertainty}, we draw $N_{MC}$ Monte Carlo samples to estimate the predictive mean $\hat{\mu}_{t+1}$ and variance $\hat{\sigma}_{t+1}^2$ by
\begin{subequations}
\begin{align}
\label{eq:mc_test}
\hat{\mu}_{t+1} &= \frac{1}{N_{MC}}  \sum_{j=1}^{N_{MC}} \mu^j(x_t), \\
%\hat{\sigma}^2_{t+1} &= \frac{1}{TN_{MC}} \sum_j^{j=N_{MC}}\sigma^j(x_t)^2, 
\hat{\sigma}^2_{t+1} &= \frac{1}{T} \sum_{j=1}^{N_{MC}}\frac{\sigma^j(x_t)^2}{N_{MC}} + \sum_{j=1}^{N_{MC}}\frac{\big(\mu^j(x_t)- \hat{\mu}_{t+1}\big)^2}{N_{MC}-1},
\end{align}
\end{subequations}
where $\hat{\mu}_{t+1}$ is used as the point prediction of the proposed approach.

%%==================
\section{Experiments}
\label{sec:experiment}
%\todo[inline]{Briefly introduce this section here.}
To compare the performance of DeepSTUQ with other state-of-the-arts, extensive experiments are conducted on real-world datasets in terms of point prediction, uncertainty quantification, and ablation study. 

%\task[(06-21) Also need brief introduction to each of the test. What is tested.]{}

%%==================
\subsection{Datasets}
Four different public datasets collected from the Caltrans Performance Measurement System (PEMS), i.e., PEMS03, PEMS04, PEMS07, and PEMS08 \cite{song2020spatial} are used for evaluation. 
%The numbers of the sensors of PEMS03, PEMS04, PEMS07 and PEMS08 are $358$, $307$, $883$ and $170$, respectively. 
The traffic flow data aggregated to $5$ minutes. 
For prediction, one-hour historic data ($12$ data points) is utilized to predict the next's ($12$ data points). 
All the datasets are split into three parts with ratio $6: 2: 2$ for training, validation/calibration, and testing, respectively. 
Table \ref{table:datasets} summarizes the statistic of the four datasets.  

\begin{table}[h]
\centering
\caption{Dataset statistic.}
\label{table:datasets}
\begin{tabular}{c|ccccccccccc}
\toprule
Dataset  & \# of Nodes & \# of Edges &\# of Steps &\\
\midrule                            
PEMS03    &$358$ &$547$ &$26,208$ \\ 
\midrule                            
PEMS04    &$307$ &$340$ &$16,992$ \\
\midrule                            
PEMS07    &$883$ &$866$ &$28,224$ \\ 
\midrule                            
PEMS08    &$170$ &$295$ &$17,856$ \\ 
\bottomrule                           
\end{tabular}
\end{table}

%%==================
\subsection{Settings}
%The setup of our approach is described as follows.
\textbf{Pre-training}.
The total number of training epochs is $100$. 
The optimizer is Adam with learning rate $0.003$ and weight decay $10^{-6}$. 
The batch size is $64$.
The relative weight $\lambda$ in Equation \eqref{eq:heter_loss} for computing the aleatoric uncertainty is $0.1$. 
The dropout rates of the graph convolutional operations in the encoder are 
%\task[(06-21) Then you need to provide a reason why two values are used for different datasets.]
$0.1$ for PEMS03, PEMS04, and PEMS07 (the adjacency matrice are relatively large), and $0.05$ for PEMS08 (the adjacency matrix is relatively small). 
The dropout rate at the final dropout layer in the decoder for all the datasets is $0.2$.

\textbf{AWA Re-training}. 
The optimizer of the AWA re-training process is Adam, and the maximum and minimum learning rates are $0.003$ and $0.00003$, respectively. 
The total number of re-training epochs is $20$, which means that $10$ models are averaged.

\textbf{Model Calibration}.
The number of Monte Carlo samples for calculating $\sigma^2$ is $10$. 
The steps and numbers of iterations of the L-BFGS optimizer are $0.02$ and $500$, respectively.

\textbf{Inference}.
To balance the inference time and model performance, we generate $10$ Monte Carlo samples for Bayesian approximation.

%%==================
\subsection{Baselines}
To compare the proposed DeepSTUQ with state-of-that-arts on point prediction and uncertainty quantification, two groups of recent traffic prediction methods are adopted as the baselines, respectively.

\subsubsection{Point Prediction Baselines}
\begin{itemize}
    \item \textbf{DCRNN} \cite{li2018diffusion} adopts diffusion convolution and sequence-to-sequence learning;
    \item GraphWaveNet (\textbf{GWN}) \cite{wu2019graph} adopts a self-adaptive adjacency matrix and dilated casual convolution.
    \item \textbf{ST-GCN} \cite{yu2018spatio} utilizes a GNN and a GCNN to forecast traffic;
    \item \textbf{ASTGCN} \cite{guo2019attention} employs Attention mechanism to model spatio-temporal dependency;
    \item \textbf{STSGCN} \cite{song2020spatial} forecasts traffic by synchronously extracting spatial-temporal correlations;
    \item \textbf{STFGNN} \cite{li2021spatial} employs a spatial-temporal fusion module and a gated dilated CNN;
    \item \textbf{AGCRN} \cite{bai2020adaptive} leverages a Node Adaptive Parameter Learning module and a Data Adaptive Graph Generation module to enhance traffic prediction performance;
    \item \textbf{DeepSTUQ/S} refers to the proposed method with single deterministic forward pass (dropout is turned off at test time).   
\end{itemize}

%\task[(06-21) In addition to what? This reads strange.]{In addition}, 
%\weizhu{Moreover,}

\subsubsection{Uncertainty Quantification Baselines}
Representative approaches of different uncertainty estimation paradigms (namely, frequentist, quantile prediction, Bayesian, and ensembling) are used as the baselines.
Note that all the following methods employ the same base model structure for fair comparison.

\begin{itemize}  
    \item \textbf{Point} prediction refers to the AGCRN model which is used here to compare with other uncertainty quantification methods;
    
    \item \textbf{Quantile} regression \cite{koenker2001quantile} is a distribution-free method which directly computes the mean, lower and upper bounds using the corresponding quantile ($0.025$, $0.5$, $0.975$);
    
    \item Mean Variance Estimation (\textbf{MVE}) \cite{nix1994estimating} refers to the method that estimates heterogeneous aleatoric uncertainty through computing Equation \eqref{eq:heter_loss};
    
    \item Monte Carlo dropout (\textbf{MCDO}) \cite{gal2016dropout} performs dropout at both training and test time, the number of Monte Carlo samples for inference is $10$;
    
    \item \textbf{Combined} refers to the method that calculates both epistemic and aleatoric uncertainty using Equation \eqref{eq:combined_loss} \cite{kendall2017uncertainties}, the number of Monte Carlo samples for inference is $10$;
    
    \item Temperature Scaling (\textbf{TS}) \cite{guo2017calibration} calibrates the aleatoric uncertainty obtained by MVE;
    
    \item Fast Geometric Ensembling (\textbf{FGE}) \cite{garipov2018loss} performs fast ensembling via varying the learning rate, the number of the stored trained models is $10$;
    
    \item Locally Weighted \textbf{Conformal} Inference \cite{lei2018distribution}, \cite{angelopoulos2021gentle} calibrates the aleatoric uncertainty obtained by MVE via conformalization;
    
    \item Conformal Forecasting Recurrent Neual Network (\textbf{CFRNN}) \cite{stankeviciute2021conformal} computes the multi-horizon uncertainty using conformal prediction;
    
\end{itemize}

\begin{table}[h]
\centering
\caption{Description of uncertainty quantification methods.}
\label{table:uq_baselines}
\begin{tabular}{c|ccccccccccc}
\toprule
Method  & Paradigm & Uncertainty Type\\
\midrule                            
Point    &deterministic      &no \\ 
\midrule                            
Quantile         &distribution-free  &aleatoric\\ 
\midrule                            
MVE              &frequentist    &aleatoric \\ 
\midrule                            
MCDO             &Bayesian  &epistemic \\ 
\midrule                            
Combined         &Bayesian  &aleatoric + epistemic \\ 
\midrule                            
TS               &frequentist  &aleatoric \\ 
\midrule                            
FGE              &ensembling  &epistemic \\ 
\midrule                            
Conformal        &frequentist  &aleatoric \\ 
\midrule                            
CFRNN            &distribution-free   &aleatoric \\ 
\midrule                            
DeepSTUQ         &Bayesian + ensembling   &aleatoric + epistemic \\ 
\bottomrule                           
\end{tabular}
\end{table}

Table \ref{table:uq_baselines} summarizes the characteristics of the uncertainty quantification methods.

%%==================
\subsection{Metrics}
Two groups of metrics are employed to evaluate the point prediction and uncertainty quantification performance, respectively.  

\subsubsection{Point Prediction Metrics}
The point traffic forecasting performance are evaluated by the following metrics.
%\task[(06-21) Consider use itemize or enumerate environment.]
%\begin{itemize}
\begin{enumerate}[label=(\alph*)]
\item Root Mean Squared Error (\textbf{RMSE}):
\begin{align}
\label{eq:rmse}
\mathrm{RMSE} = \sqrt{ \frac{1}{N}\sum_{i=1}^N(\hat{y}_i-y_i)^2},
\end{align} 
%\task[(06-21) For RMSE, MAE, ..., MPIW in these equations, use mathrm or even text\{\}.]{}
where $y_i$ is the ground truth, and $\hat{y}_i$ is the prediction. 

\item Mean Absolute Error (\textbf{MAE}): 
\begin{align}
\label{eq:mae}
\mathrm{MAE} = \frac{1}{N}\sum_{i=1}^N \left |\hat{y}_i-y_i \right|.
\end{align} 

\item 
Mean Absolute Percentage Error (\textbf{MAPE}): 
\begin{align}
\label{eq:mape}
\mathrm{MAPE} = \frac{1}{N}\sum_{i=1}^N\left |\frac{\hat{y}_i-y_i}{y_i} \right|.
\end{align} 

\end{enumerate}

\subsubsection{Uncertainty Quantification Metrics}
The uncertainty quantification performance are evaluated by the following metrics.
%\task[(06-21) Ditto]{(i).} 

\begin{enumerate}[label=(\alph*)]
\item Mean Negative Log-Likelihood (\textbf{MNLL}): 
\begin{align}
\label{eq:mnnl}
\mathrm{MNLL} = \frac{1}{N}\sum_{i=1}^N- \log \mathcal{N}(y_i;\hat{\mu}_i,\hat{\sigma}^2_i),
\end{align} 
where $\hat{\mu}_i$ and $\hat{\sigma}^2_i$ are the predicted mean and predicted variance, respectively.

\item  Prediction Interval Coverage Probability (\textbf{PICP}).
The predicted lower and upper bounds of the prediction interval are denoted by $\hat{y_L}$ and $\hat{y_U}$, respectively. 
%\task[(06-21) Break into small pieces.]
%\weizhu{If the prediction interval is set to $95\%$, it means that the expected probability of a ground truth data point falling into the range %$[\hat{y_L}$, $\hat{y_U}]$ is $95\%$.
%Accordingly, $\hat{y_U}_i=\hat{\mu}_i+1.96 \hat{\sigma}_i$, and $\hat{y_L}_i=\hat{\mu}_i-1.96 \hat{\sigma}_i$. }
Let the significance level $\alpha$ be $5\%$, which means that the expected probability of a ground truth data point falling into the range $[\hat{y}_L$, $\hat{y}_U]$ is $95\%$ ($100\% - \alpha = 95\%$).
Accordingly, under Gaussianity assumption, $\hat{y}_{U_i}=\hat{\mu}_i+1.96 \hat{\sigma}_i$, and $\hat{y}_{L_i}=\hat{\mu}_i-1.96 \hat{\sigma}_i$.
Let {$k_i^j$} indicate whether the real speed value of a road segment $j$ at time $i$ is captured by the estimated prediction interval, and we have
\begin{align}
\label{eq:picp1}
k_i = \begin{cases}
    1, \text{ if } \hat{y}_{L_i} \leq y_i \leq \hat{y}_{U_i}\\
    0, \text{ else}. 
\end{cases}
\end{align} 
Then PICP can be formulated by
\begin{align}
\label{eq:picp3}
\mathrm{PICP} = \frac{1}{N} \sum_{i=1}^N k_i.
\end{align} 
%
%Let the significant level $\alpha$ be $5\%$, accordingly the ideal PICP should be equal or greater than $95\%$ ($100\% - \alpha = 95\%$).
Ideally, PICP should be equal or greater than $95\%$.

\item  Mean Prediction Interval Width (\textbf{MPIW}):
\begin{align}
\label{eq:mpiw}
\mathrm{MPIW} = \frac{1}{N}\sum_{i=1}^N\hat{y}_{U_i} - \hat{y}_{L_i}.
\end{align} 
\end{enumerate}

%%==================
\subsection{Point Prediction Results}
\begin{table*}[htbp]
%\small
\centering
\caption{Point prediction results on PEMS03, PEMS04, PEMS07, and PEMS08, where best and second best results are highlighted in bold and underlined, respectively.}
\label{table:point_results}
\begin{tabular}{c|ccccccccccc}
\toprule
Dataset  & Metrics & DCRNN  &ST-GCN &GWN  &ASTGCN  &STSGCN &STFGNN &AGCRN  &DeepSTUQ/S &DeepSTUQ\\
                           
\midrule
\multirow{3}{*}{PEMS03}    &MAE             &$18.18$  &$17.49$  &$19.85$    &$17.69$   &$17.48$  &$16.77$  &$16.05$  &\underline{$15.38$}  &\pmb{$15.13$} \\  
                           &RMSE            &$30.31$  &$30.12$  &$32.94$    &$29.66$   &$29.21$  &$28.34$  &$28.61$  &\underline{$27.03$}  &\pmb{$26.77$} \\ 
                           &MAPE ($\%$)     &$18.91$  &$17.15$  &$19.31$    &$19.40$   &$16.78$  &$16.30$  &$15.19$  &\underline{$14.45$}  &\pmb{$14.03$}\\ 
\midrule
\multirow{3}{*}{PEMS04}    &MAE             &$24.70$  &$22.70$  &$24.14$    &$22.93$   &$21.19$  &$19.83$  &$19.83$  &\underline{$19.42$}  &\pmb{$19.11$} \\  
                           &RMSE            &$38.12$  &$35.55$  &$37.60$    &$35.22$   &$33.65$  &$31.88$  &$32.26$  &\underline{$32.07$}  &\pmb{$31.68$} \\ 
                           &MAPE ($\%$)     &$17.12$  &$14.59$  &$17.93$    &$16.56$   &$13.90$  &$13.02$  &\underline{$12.97$}  &$12.98$  &\pmb{$12.71$} \\ 
                           
\midrule
\multirow{3}{*}{PEMS07}    &MAE             &$25.30$  &$25.38$  &$26.85$  &$28.05$   &$24.26$   &$22.07$  &$20.94$   &\underline{$20.76$} &\pmb{$20.36$} \\  
                           &RMSE            &$35.58$  &$38.78$  &$42.78$  &$42.57$   &$39.03$   &$35.80$  &$34.98$   &\underline{$34.12$} &\pmb{$33.71$} \\  
                           &MAPE ($\%$)     &$11.66$  &$11.08$  &$12.12$  &$19.32$   &$10.21$   &$9.21$   &\underline{$8.85$}    &$8.90$ &\pmb{$8.63$} \\
                           
\midrule                           
\multirow{3}{*}{PEMS08}    &MAE             &$17.86$  &$18.02$  &$19.13$  &$18.61$   &$17.13$  &$16.64$  &$15.95$   &\underline{$15.74$} &\pmb{$15.44$} \\  
                           &RMSE            &$27.83$  &$27.83$  &$31.05$  &$28.16$   &$26.80$  &$26.22$  &$25.22$   &\underline{$24.93$} &\pmb{$24.60$} \\   
                           &MAPE ($\%$)     &$11.45$  &$11.40$  &$12.68$  &$13.08$   &$10.96$  &$10.60$  &\underline{$10.09$}   &$10.31$ &\pmb{$10.06$} \\  
\bottomrule
\end{tabular}
\end{table*}

\begin{figure*}[!t]
\centering
\subfloat[]{\includegraphics[width=0.33\textwidth]{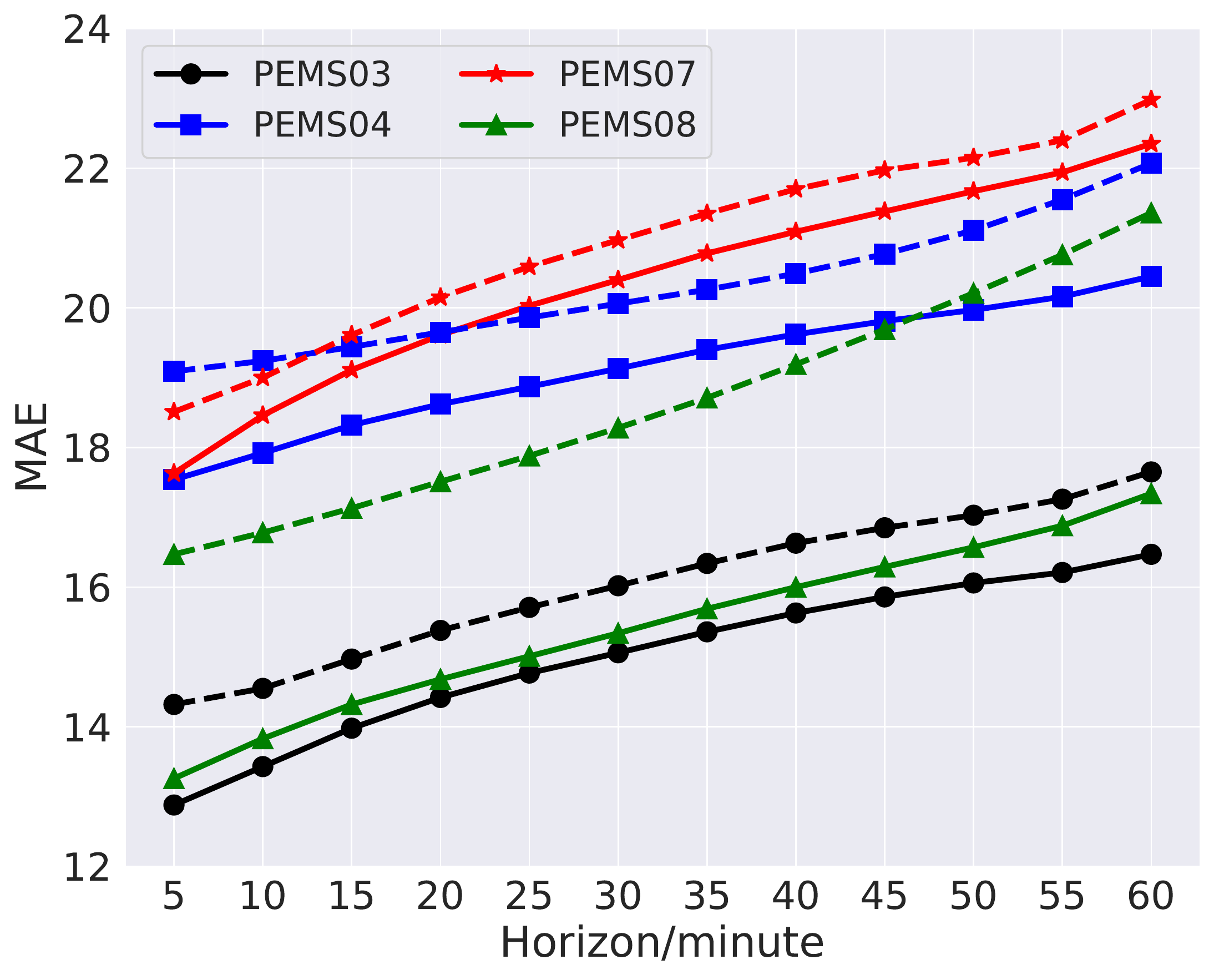}%
\label{fig:mae}}
%\\
%\hfil
\subfloat[]{\includegraphics[width=0.34\textwidth]{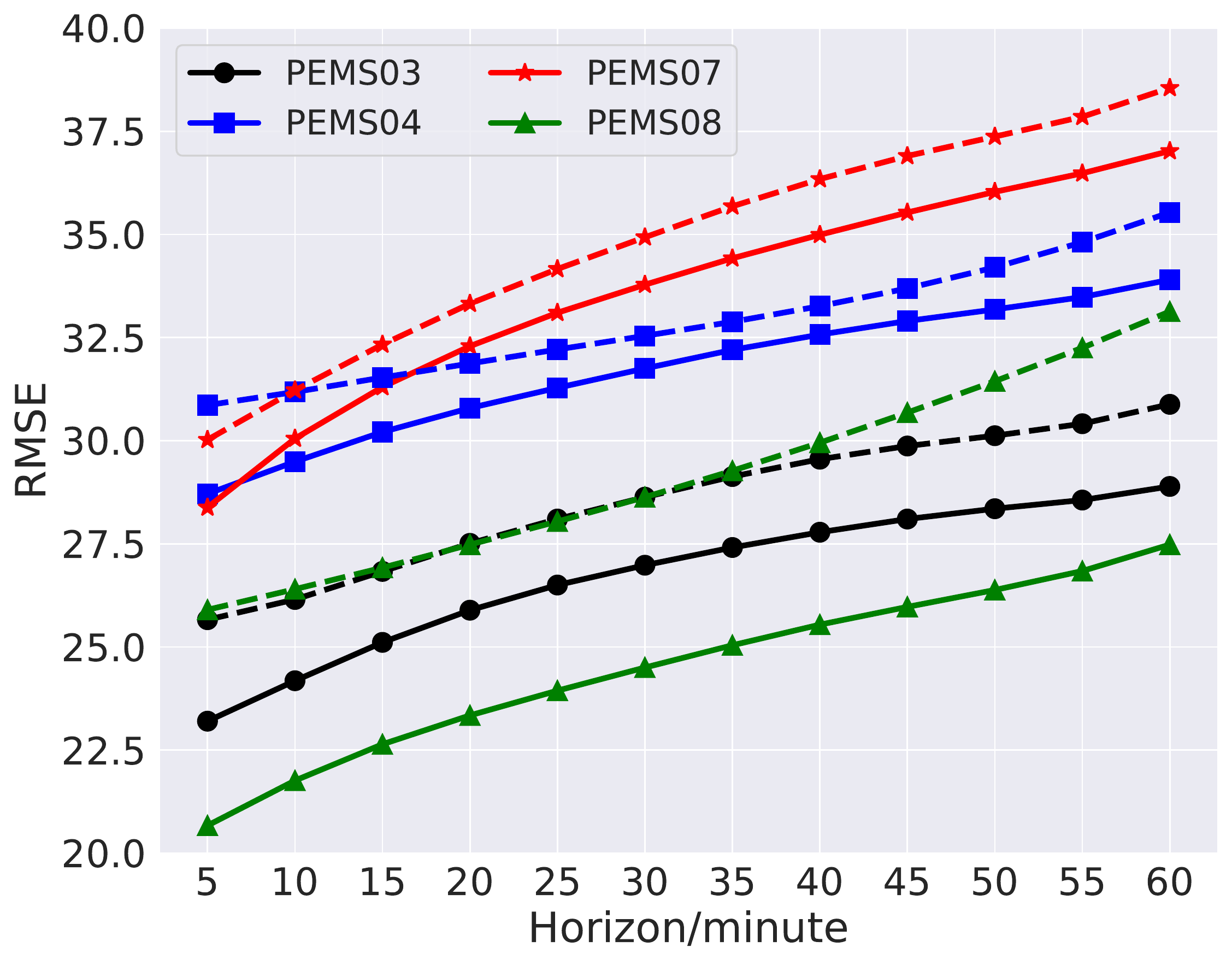}%
\label{fig:rmse}}
%\\
%\hfil
\subfloat[]{\includegraphics[width=0.33\textwidth]{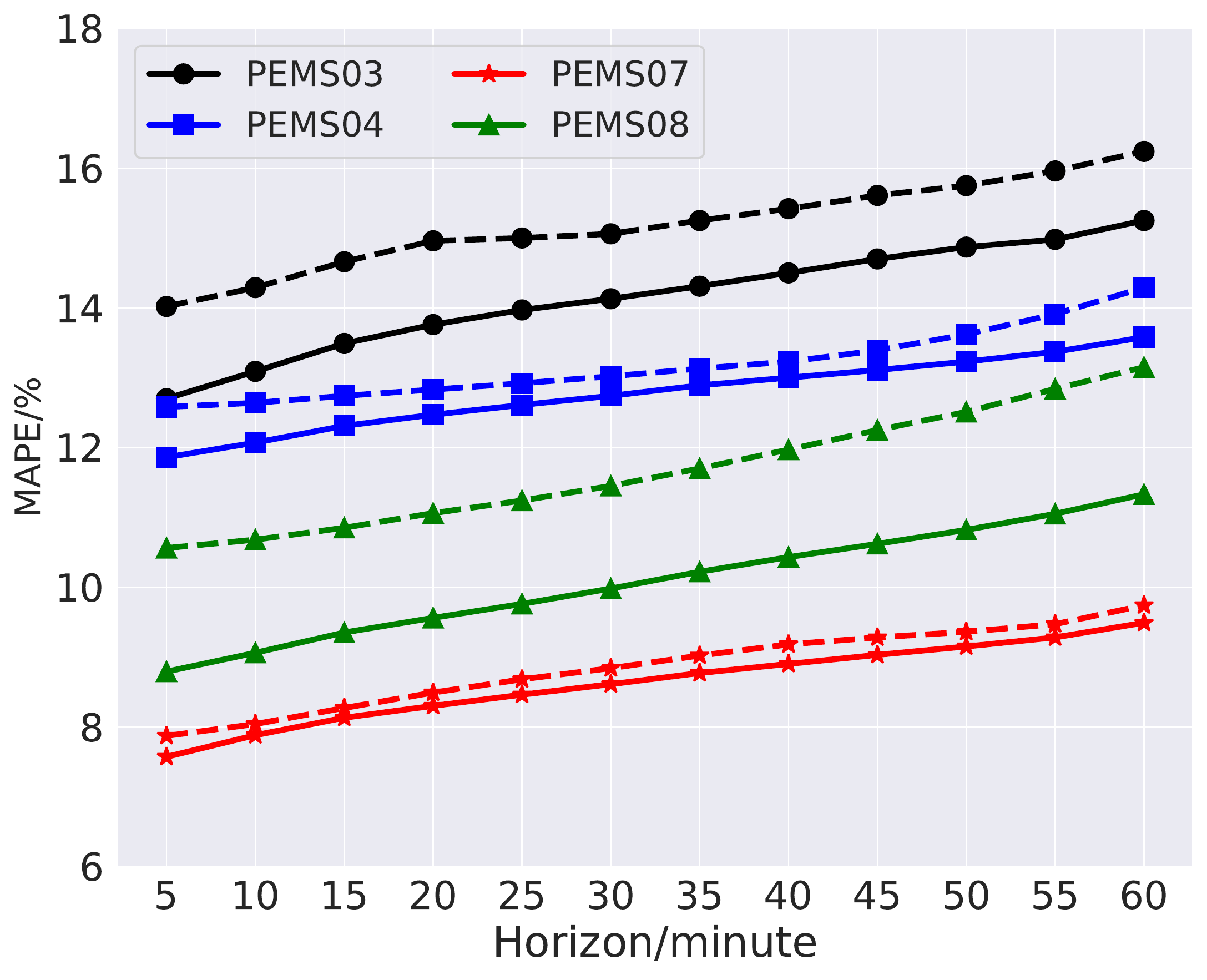}%
\label{fig:mape}}
\caption{Point prediction results with respect to various forecast horizons, where solid and dashed lines denote DeepSTUQ and AGCRN, respectively. (a) MAE. (b) RMSE. (c) MAPE.}
\label{fig:horizon_results}
\end{figure*}

The point prediction results of DeepSTUQ are compared with the aforementioned state-of-the-art methods first for performance evaluation.   
The obtained point prediction results are demonstrated in Table \ref{table:point_results}.  
As it can be seen from the results, with only $10$ Monte Carlo samples, DeepSTUQ achieves the smallest RMSEs, MAEs, and MAPEs, which suggests that DeepSTUQ has the best performance on point traffic flow prediction.  
In addition, the proposed method\,---\,even with only one single deterministic forward pass, namely DeepSTUQ/S\,---\,also outperforms other state-of-the-art methods, which indicates that the proposed method is competitive on point prediction at nearly the same inference time cost as other deterministic approaches.     
This is because that variational inference can obtain a set of solutions around on one local minimum, and deep ensembling can find multiple local minimums in the solution space. By combining these two approaches, DeepSTUQ is capable of finding better sub-optimal solutions and have better generalization ability compared to deterministic methods, and consequently has better performance regarding point prediction.
Fig. \ref{fig:horizon_results} shows the point prediction performance with respect to different horizons, which suggests that DeepSTUP has better performance than AGCRN at each time step for all the datasets.
%\task[(06-21) Insights. Why good? This is your first experiment, reader expects a lot of analysis and hypothesis.]{}

%\task[(06-21) Is this good? If so, why? If not, why present?]{the point prediction performance decreases} as the prediction horizon increase. 

%%==================
\subsection{Uncertainty Quantification Results}

\begin{table*}[htbp]
\footnotesize
\centering
\caption{Uncertainty quantification results on PEMS03, PEMS04, PEMS07, and PEMS08, the best results are highlighted in bold. The results are evaluated according to the following criteria. % \cite{pearce2018high}.
Any PICP $\geq 95\%$ is the best and the smallest corresponding MPIW is the best.
If all the PICP $ < 95\%$, then the largest PICP is the best. MPIW only assessed when the corresponding PICP $\geq 95\%$.}
\label{table:uq_results}
\begin{tabular}{c|cccccccccccc}
\toprule
Dataset  & Metrics &Point &Quantile  &MVE &MCDO &Combined &TS  &FGE &Conformal &CFRNN &DeepSTUQ\\
\midrule                       
\multirow{6}{*}{PEMS03}    &MAE           &$16.05$ &$16.06$  &$15.97$   &$15.23$   &$15.29$   &$15.97$  &$15.23$   &$15.97$  &$16.05$ &\pmb{$15.13$} \\ 
                           &RMSE          &$28.61$ &$28.40$  &$28.17$   &$26.95$   &$27.13$   &$28.17$  &$26.99$  &$28.17$  &$28.61$ &\pmb{$26.77$} \\ 
                           &MAPE($\%$)    &$15.19$ &$15.50$  &$15.08$   &$14.39$   &$14.60$   &$15.08$  &$14.36$ &$15.08$  &$15.19$ &\pmb{$14.03$}\\ 
                        
                           &MNLL          &$-$  &$-$ &$3.53$    &$12.32$   &$3.39$     &$3.49$           &$25.94$   &$3.53$  &$-$    &\pmb{$3.38$} \\ 
                           &PICP($\%$)    &$-$  &$89.49$ &$92.06$   &$43.92$   &$93.64$    &$93.51$     &$31.15$ &$93.21$  &$93.00$  &\pmb{$94.75$} \\ 
                           &MPIW          &$-$  &$65.60$ &$74.04$   &$19.73$   &$73.26$    &$79.79$     &$12.81$ &$76.72$  &$82.79$  &$76.91$ \\

\midrule 
\multirow{6}{*}{PEMS04}    &MAE        &$19.83$ &$20.08$   &$19.86$   &$19.15$       &$19.23$     &$19.86$  &\pmb{$19.08$} &$19.86$   &$19.83$  &$19.11$ \\ 
                           &RMSE       &$32.26$ &$32.76$   &$32.30$   &\pmb{$31.49$} &$31.73$    &$32.30$  &$31.59$ &$32.30$   &$32.26$         &$31.68$   \\ 
                           &MAPE($\%$) &$12.97$ &$13.06$   &$13.25$   &$12.77$       &$12.87$     &$12.97$  &\pmb{$12.69$} &$13.25$   &$12.97$    &$12.71$\\ 
                           
                           &MNLL       &$-$     &$-$ &$3.71$    &$23.17$     &$3.63$     &$3.70$              &$15.47$ &$3.71$  &$-$ &\pmb{$3.57$} \\ 
                           &PICP($\%$) &$-$     &$91.87$ &$93.10$   &$34.18$     &\pmb{$95.16$}    &$94.86$ &$42.60$   &$94.33$ &$94.65$    &\pmb{$95.23$}  \\ 
                           &MPIW       &$-$     &$91.72$ &$102.97$   &$17.30$      &$108.44$    &$115.48$   &$21.95$   &$109.74$  &$118.83$   &\pmb{$105.42$} \\ 
                          
\midrule     
\multirow{6}{*}{PEMS07}    &MAE         &$20.94$ &$21.28$ &$21.07$ &$20.61$ &$20.37$ &$21.07$       &$20.42$  &$21.07$  &$20.94$ &\pmb{$20.36$} \\ 
                           &RMSE        &$34.98$ &$35.76$ &$34.94$ &$34.20$ &\pmb{$33.64$} &$34.94$    &$34.13$ &$34.94$  &$34.98$  &$33.71$\\ 
                           &MAPE($\%$)  &$8.85$  &$8.95$ &$8.88$  &$8.73$  &$8.68$ &$8.88$     &$8.622$ &$8.88$  &$8.85$  &\pmb{$8.63$} \\ 
                           
                           &MNLL        &$-$    &$-$ &$3.80$  &$9.88$  &\pmb{$3.60$} &$3.78$    &$22.31$  &$3.80$  &$-$  &\pmb{$3.60$}\\ 
                           &PICP($\%$)  &$-$    &$91.83$ &$93.86$ &$53.74$ &\pmb{$95.80$} &\pmb{$95.53$}    &$38.05$ &$94.78$  &$94.65$ &\pmb{$95.74$} \\  
                           &MPIW        &$-$    &$96.63$ &$112.99$ &$31.16$ &$112.36$ &$127.00$   &$19.03$ &$118.79$  &$118.83$ &\pmb{$111.68$}\\

\midrule 
\multirow{6}{*}{PEMS08}    &MAE        &$15.95$  &$16.40$ &$16.29$   &$15.87$         &$15.51$   &$16.29$     &$15.79$    &$16.29$  &$15.95$   &\pmb{$15.44$} \\ 
                           &RMSE       &$25.22$  &$25.79$ &$25.71$   &$25.05$         &$24.64$   &$25.71$     &$24.96$   &$25.71$  &$25.22$   &\pmb{$24.60$} \\ 
                           &MAPE($\%$) &$10.09$  &$10.56$ &$10.36$   &\pmb{$10.05$}   &$10.14$   &$10.36$   &$10.17$    &$10.36$  &$10.09$   &$10.06$  \\
                           
                           &MNLL       &$-$     &$-$  &$3.63$     &$11.77$   &$3.45$     &$3.62$       &$11.58$ &$3.63$  &$-$     &\pmb{$3.44$}\\ 
                           &PICP($\%$) &$-$     &$93.95$ &$94.79$    &$49.91$   &\pmb{$95.88$} &\pmb{$97.15$}     &$50.15$ &\pmb{$95.37$}  &\pmb{$95.16$} &\pmb{$95.65$}\\ 
                           &MPIW       &$-$     &$82.13$ &$93.13$    &$23.53$   &$91.45$    &$113.34$     &$23.57$  &$96.94$  &$96.34$      &\pmb{$89.63$} \\

\bottomrule                           
\end{tabular}
\end{table*}

\begin{figure*}[htbp]
\centering
\subfloat[]{\includegraphics[width=0.49\textwidth]{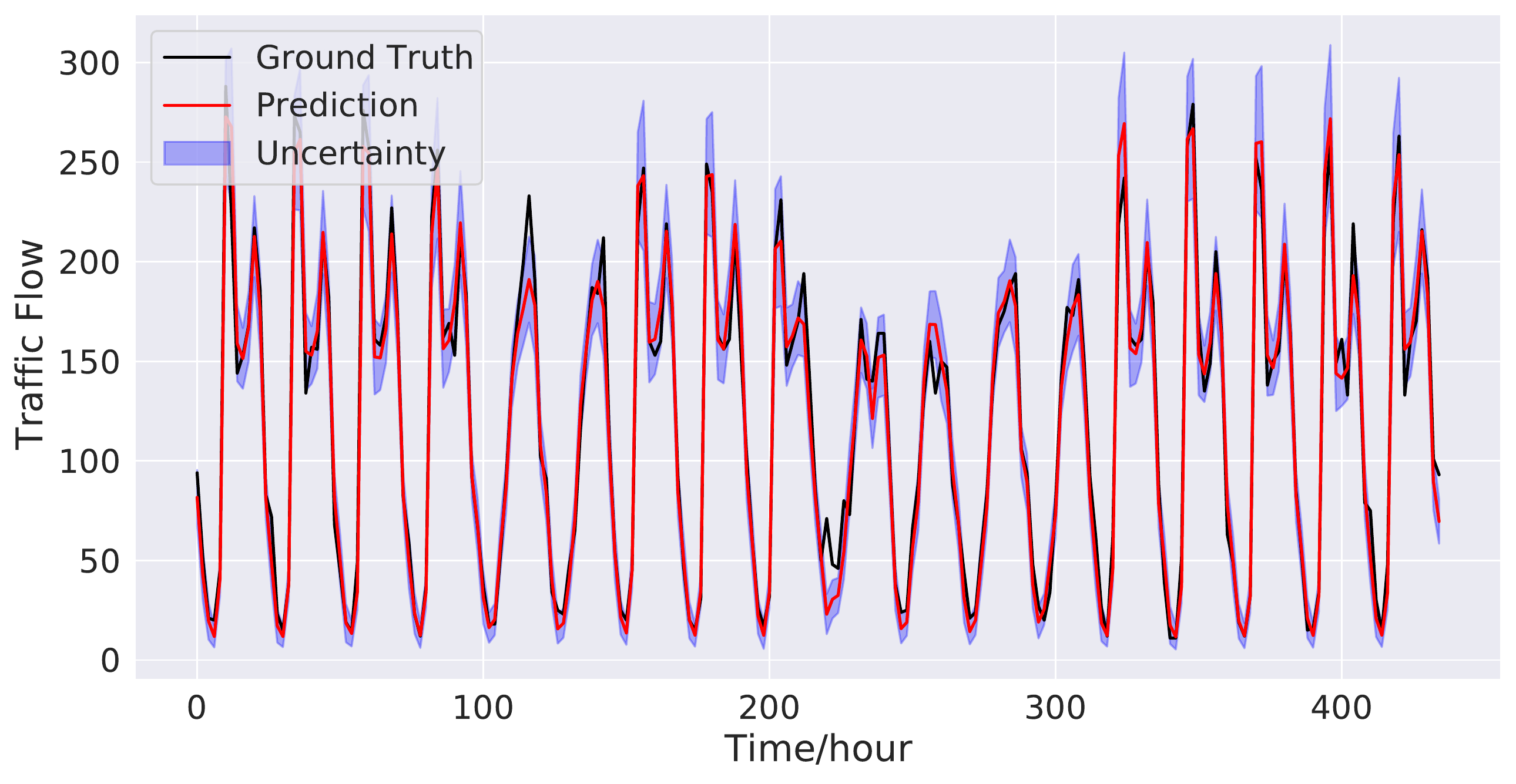}%
\label{fig:pems03_result}}
%\hfil
\subfloat[]{\includegraphics[width=0.5\textwidth]{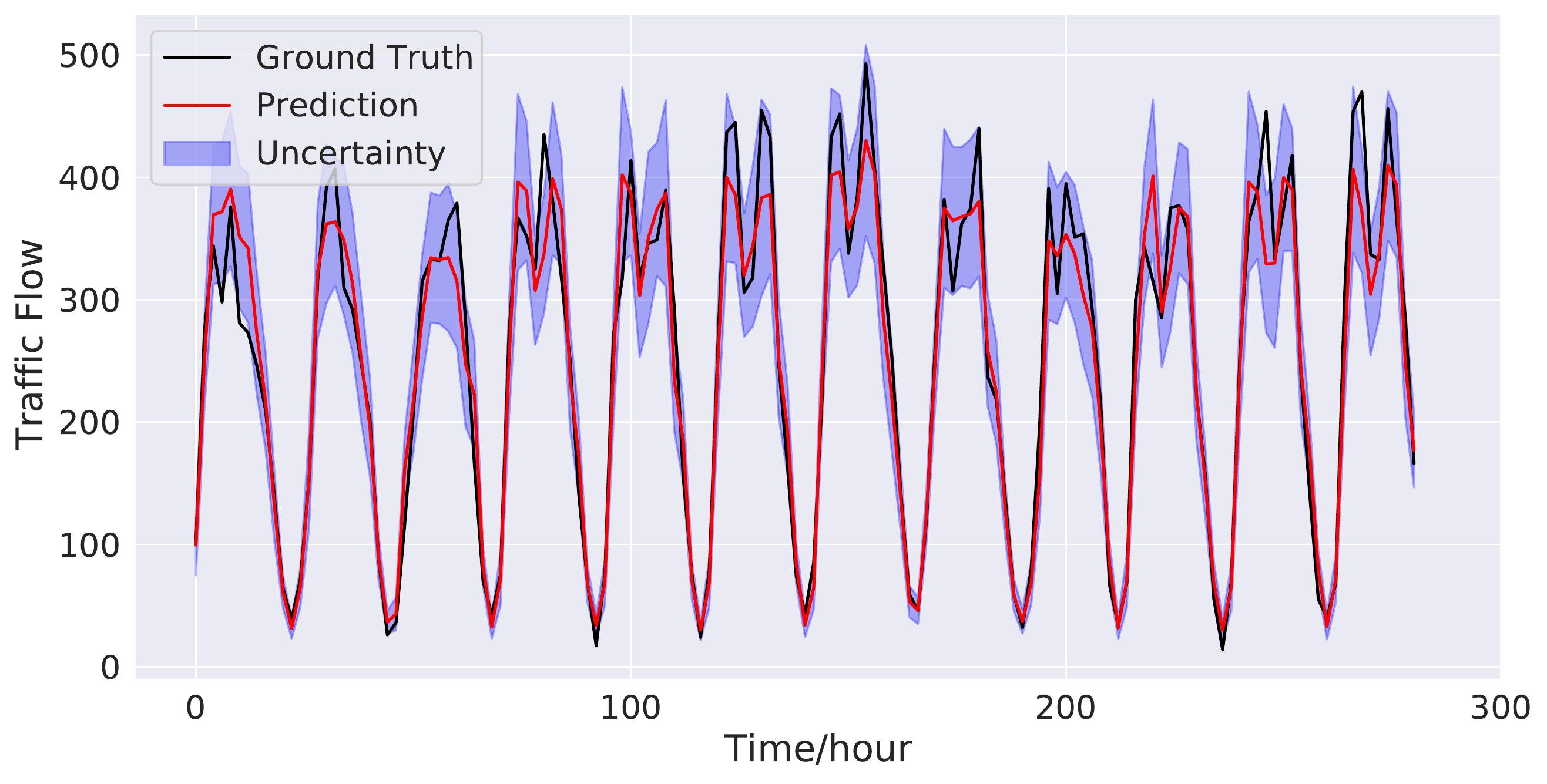}%
\label{fig:pems04_result}}
\hfil
\subfloat[]{\includegraphics[width=0.5\textwidth]{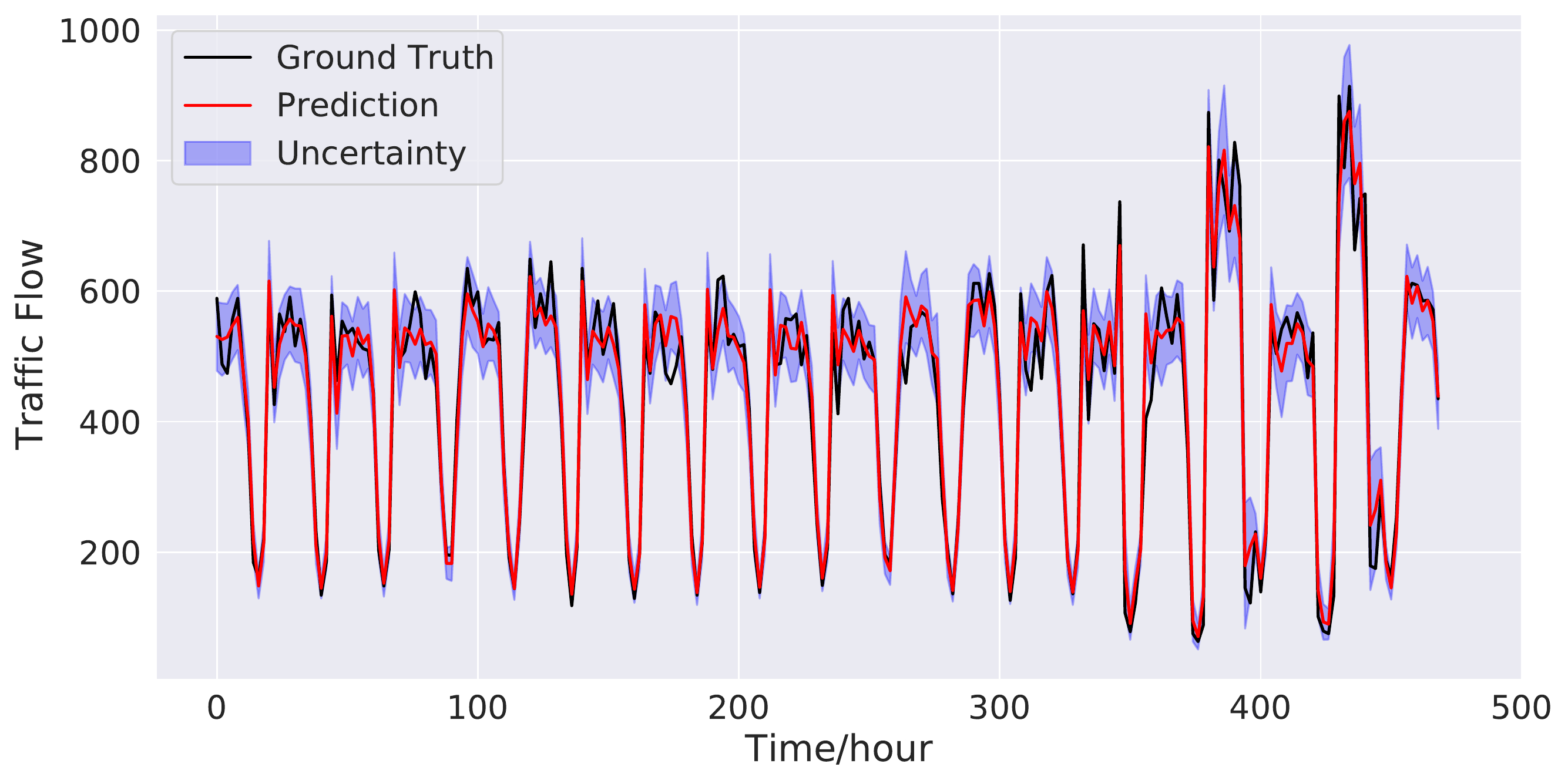}%
\label{fig:pems07_result}}
%\hfil
\subfloat[]{\includegraphics[width=0.49\textwidth]{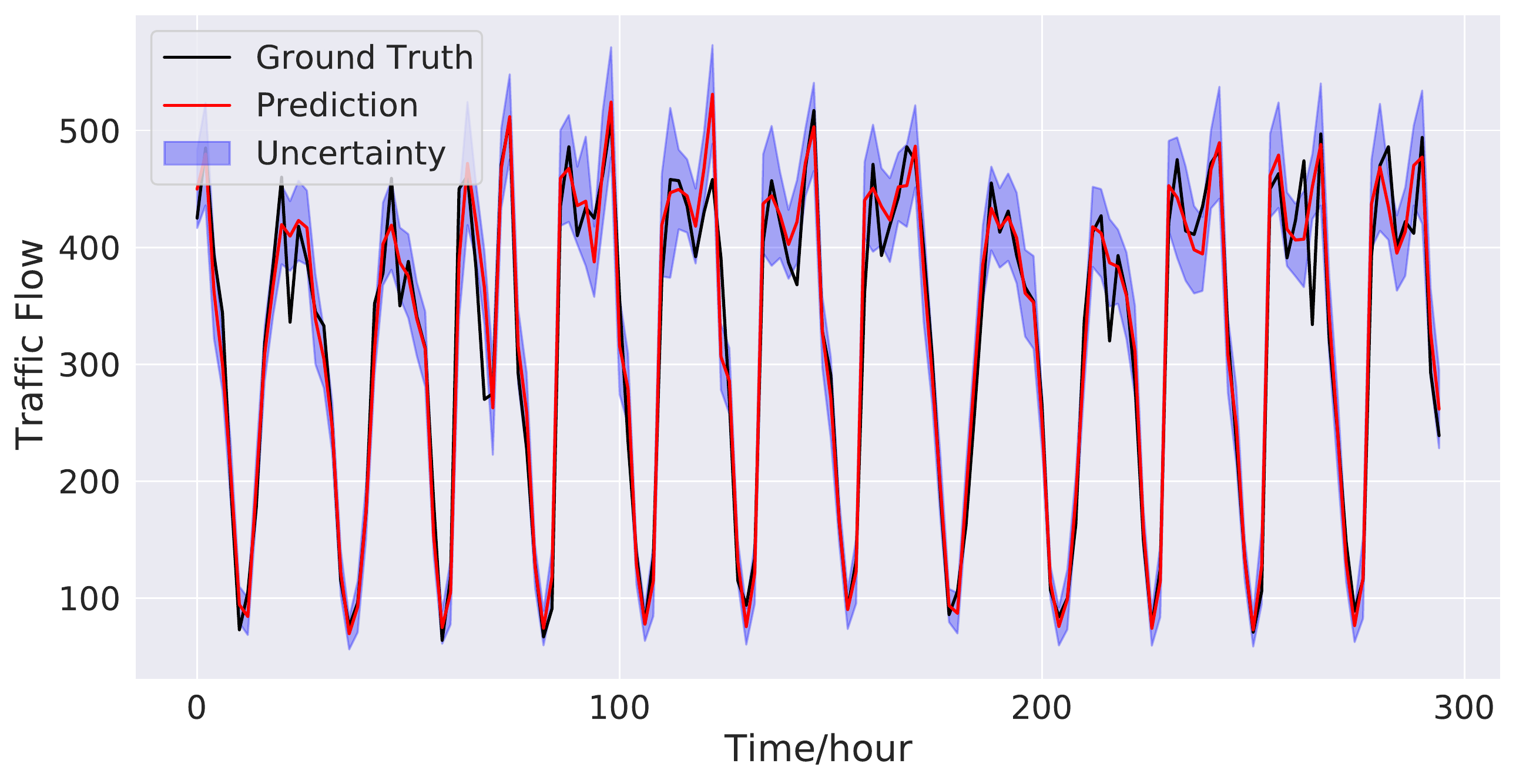}%
\label{fig:pems08_result}}
\caption{Uncertainty quantification results on randomly selected road segments from different datasets. (a) PEMS03. (b) PEMS04. (c) PEMS07. (d) PEMS08.}
\label{fig:uq_results}
\end{figure*}

\begin{figure}[htbp]
\centering
\includegraphics[width=0.42\textwidth]{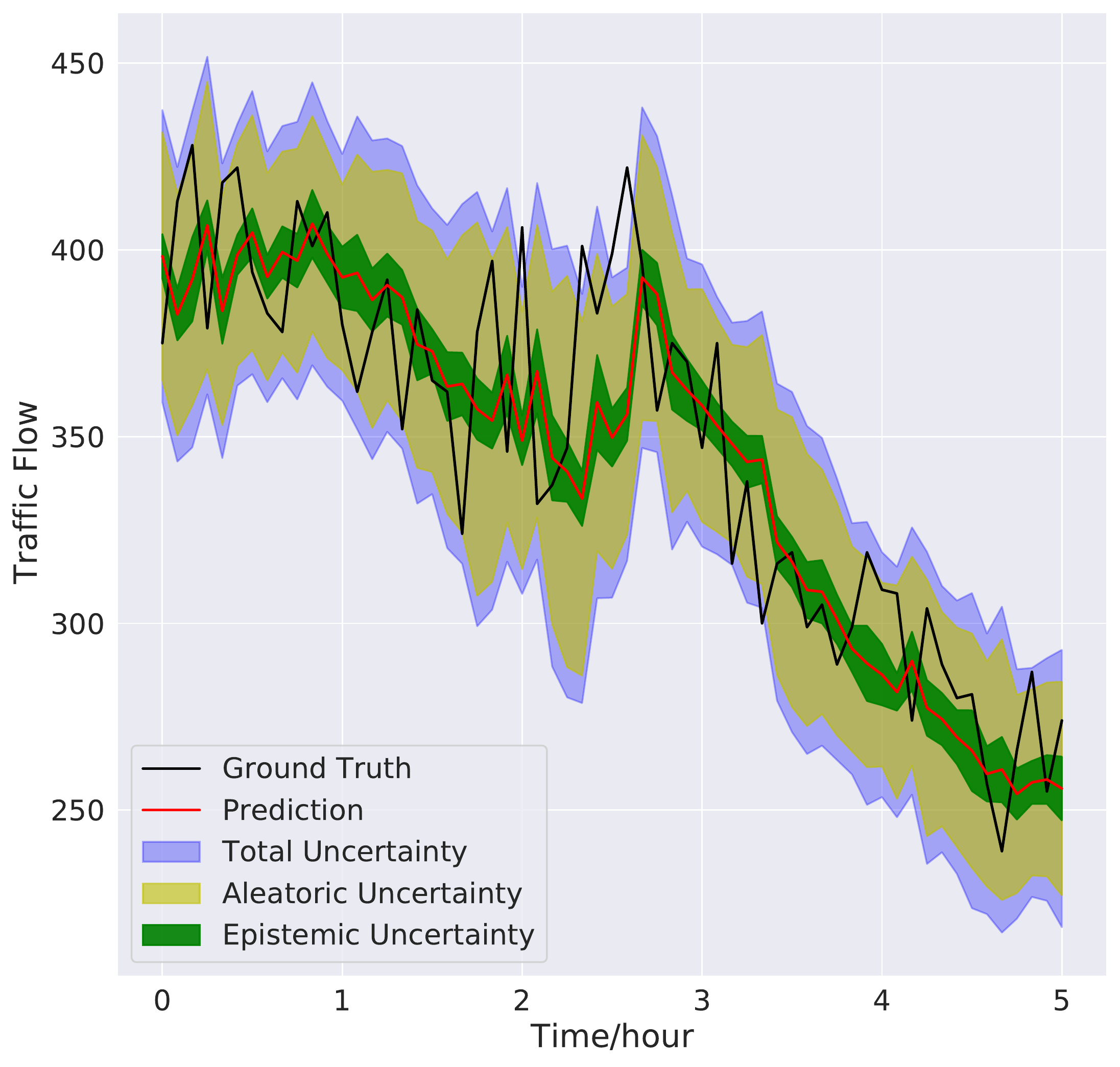}
\caption{Quantification results of different uncertainties on partial data from a randomly selected segment of PEMS08.}
\label{fig:uncertainties}
\end{figure}

\begin{figure*}[htbp]
\centering
\subfloat[]{\includegraphics[width=0.35\textwidth]{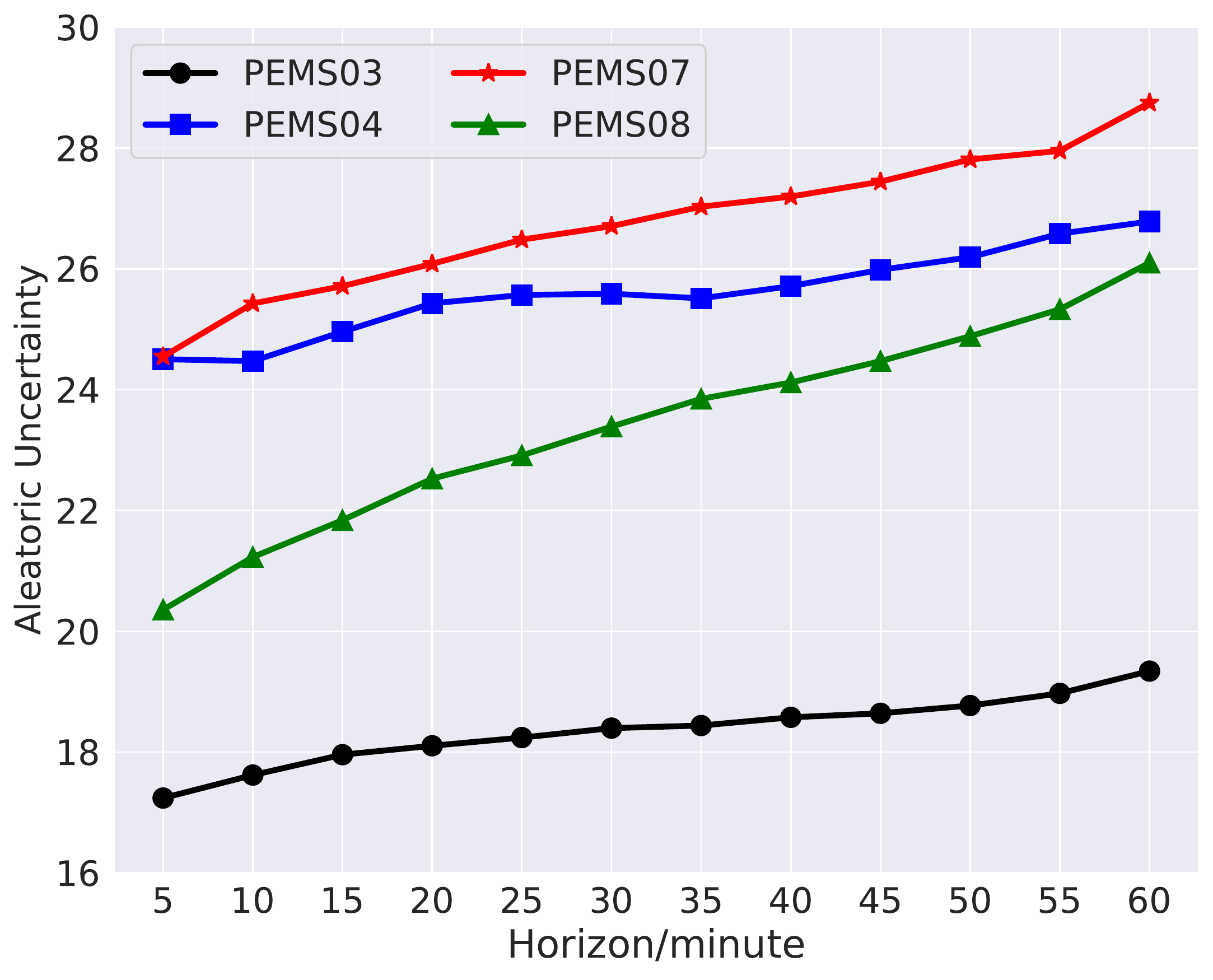}%
\label{fig:alea}}
\quad
\subfloat[]{\includegraphics[width=0.35\textwidth]{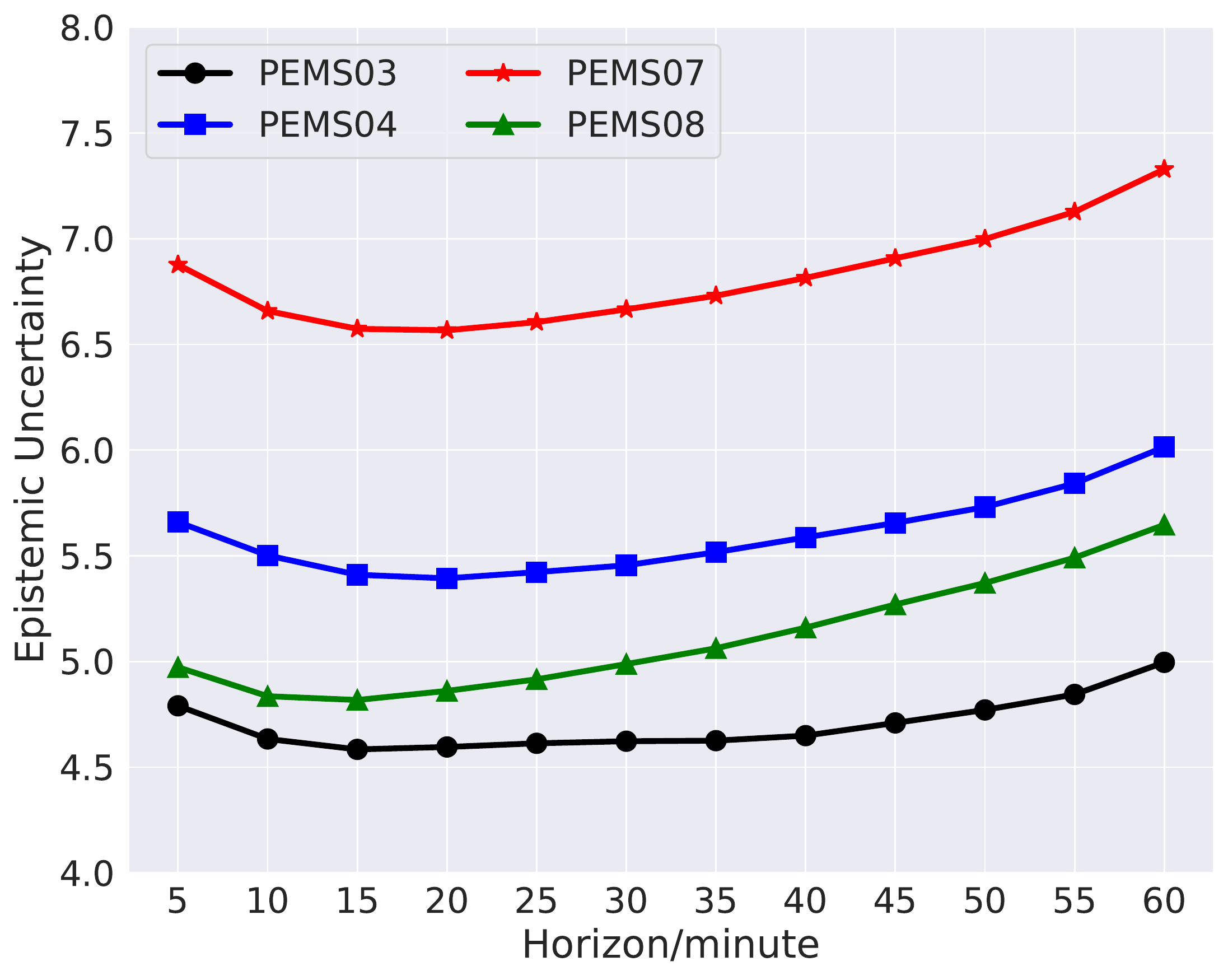}%
\label{fig:epis}}
\caption{Uncertainty quantification results with respect to different horizons. (a) Aleatoric uncertainty. (b) Epistemic uncertainty. }
\label{fig:uq_horizon}
\end{figure*}

To evaluate the uncertainty quantification performance, DeepSTUQ is compared with the uncertainty quantification baselines, whose results are demonstrated in Table \ref{table:uq_results} and Figs. \ref{fig:uq_results}--\ref{fig:uq_horizon}.  
According to the results in the table, the proposed approach has the best overall performance regarding both the point prediction and uncertainty quantification results compared with others.  
As it is observed from Fig. \ref{fig:uq_results}, DeepSTUQ can forecast traffic flow accurately and provide valid coverage for future ground truth.
%\task[(07-04) This is not very straightforward. How to define ``good'' using the figure? Fall in the shades?]{good uncertainty estimation} of the prediction.  
Fig. \ref{fig:uncertainties} illustrates that in traffic flow forecasting, the aleatoric uncertainty is much larger than the epistemic uncertainty. 
Hence, considering total uncertainty can provide better uncertainty estimation than considering either one alone.  
Fig. \ref{fig:uq_horizon} shows that, for all the datasets, generally, both aleatoric and epistemic uncertainty increase as the prediction horizons extend, which implies that short-term traffic flow forecasting is more reliable than long-term one.
%\fix[Provide two or three references.]{}
The conclusion accords with the intuition and results in the literatures \cite{li2018diffusion,bai2020adaptive,li2021spatial}

In terms of uncertainty quantification, the aleatoric uncertainty-aware approaches, i.e., MVE and TS, outperform the epistemic uncertainty-aware approaches, which suggests that the traffic uncertainty is mainly data-related.  
The results indicate that only considering epistemic uncertainty improves the estimation of the predictive mean (which results in better point estimation) but underestimates the variance significantly. 
This conclusion is supported by \cite{wu2021quantifying} as well.
Although we have made a strong Gaussianity assumption on the likelihood of the aleatoric uncertainty, the obtained experimental results indicate that the methods using this assumption (i.e., MVE, Combined, TS, and DeepSTUQ) outperform the distribution-free method, Quantile.
Additionally, the PICPs obtained by DeepSTUQ on the four datasets are very close to or larger than $95\%$, which implies that the Gaussian distribution assumption is credible.

According to the experimental results, we can also see that when only the epistemic uncertainty is considered using variational inference (MCDO) or deep ensembling (FGE), the traffic flow point prediction performance is improved compared to deterministic methods but the uncertainty quantification performance is poor.  
If merely the aleatoric uncertainty is taken into account (MVE, TS, Conformal, and CFRNN), the uncertainty quantification performance is satisfying while the point prediction slightly decreases compared to deterministic methods.   
On the other hand, if both the epistemic and aleatoric uncertainties are estimated, e.g., Combined and DeepSTUQ, the point prediction and uncertainty quantification performance are both improved.

%%==================
\subsection{Ablation Study}
Three groups of experiments are conducted to verify the effects of the proposed AWA training, the proposed model calibration method, and different numbers of Monte Carlo samples, respectively. 

%%==================
\subsubsection{Effect of AWA Re-training}

The prediction performance of the same pre-trained model prior to and following AWA post-processing re-training are compared. 
Table \ref{table:ablation_swa} demonstrates that after AWA training, the point prediction performance has improved, indicating that the proposed AWA training method can approximate the deep ensembling method using only one single model with mere $20$ additional epochs. 
Therefore, compared to conventional deep ensembling, DeepSTUQ requires less time and memory. 

\begin{table}[!t]
\centering
\caption{Ablation study results on AWA training.}
\label{table:ablation_swa}
\begin{tabular}{c|ccccccccccc}
\toprule
Dataset  & Metrics & No AWA &AWA\\
\midrule                            
\multirow{3}{*}{PEMS03}    &MAE           &$15.29$ &\pmb{$15.13$}  \\ 
                           &RMSE          &$27.13$ &\pmb{$26.77$} \\ 
                           &MAPE($\%$)    &$14.60$ &\pmb{$14.03$} \\ 

\midrule
\multirow{3}{*}{PEMS04}    &MAE        &$19.23$  &\pmb{$19.11$} \\ 
                           &RMSE       &$31.73$  &\pmb{$31.68$} \\ 
                           &MAPE($\%$) &$12.87$  &\pmb{$12.71$} \\ 
\midrule     
           
\multirow{3}{*}{PEMS07}    &MAE         &$20.37$  &\pmb{$20.36$} \\ 
                           &RMSE        &\pmb{$33.64$}  &$33.71$ \\ 
                           &MAPE($\%$)  &$8.68$   &\pmb{$8.63$} \\ 
                           
\midrule 
\multirow{3}{*}{PEMS08}    &MAE        &$15.51$ &\pmb{$15.44$}  \\ 
                           &RMSE       &$24.64$ &\pmb{$24.60$}  \\ 
                           &MAPE($\%$) &$10.14$ &\pmb{$10.06$}  \\ 
\bottomrule                           
\end{tabular}
\end{table}

%%==================
\subsubsection{Effect of Model Calibration}
The uncertainty quantification performance of the same model before and after applying the calibration method are compared. 
From the results illustrated in Tables \ref{table:ablation_ts} and \ref{table:uq_results} (results of MVE and TS), it can be seen that the uncertainty quantification performance has been further improved after model calibration, indicating that the proposed model calibration method is effective. 

\begin{table}[!t]%[htbp]
%\small
\centering
\caption{Ablation study results on model calibration.}
\label{table:ablation_ts}
\begin{tabular}{c|ccccccccccc}
\toprule
Dataset  & Metrics &No Calibration  & Calibration\\
\midrule                            
\multirow{3}{*}{PEMS03}    &MNLL        &$3.39$  &\pmb{$3.38$} \\ 
                           &PICP($\%$)  &$94.22$ &\pmb{$94.75$ }\\ 
                           %&MPIW        &$38.02$ &$39.24$ \\
                           &MPIW        &$74.51$ &$76.91$ \\

\midrule
\multirow{3}{*}{PEMS04}    &MNLL       &\pmb{$3.57$}    &\pmb{$3.57$} \\ 
                           &PICP($\%$) &$94.90$   &\pmb{$95.23$} \\ 
                           %&MPIW       &$52.73$   &\pmb{$53.79$} \\ 
                           &MPIW       &$103.35$   &\pmb{$105.42$} \\

\midrule     
\multirow{3}{*}{PEMS07}    &MNLL        &\pmb{$3.60$}  &\pmb{$3.60$}\\ 
                           &PICP($\%$)  &\pmb{$95.38$} &\pmb{$95.74$}\\  
                           %&MPIW        &\pmb{$55.54$} &$56.98$\\ 
                           &MPIW        &\pmb{$108.85$} &$111.68$\\ 
                           
\midrule 
\multirow{3}{*}{PEMS08}    &MNLL       &$3.45$  &\pmb{$3.44$}\\ 
                           &PICP($\%$) &\pmb{$96.28$} &\pmb{$95.65$}\\ 
                           %&MPIW       &$48.09$ &\pmb{$45.73$}\\ 
                           &MPIW       &$94.25$ &\pmb{$89.63$}\\ 
\bottomrule                           
\end{tabular}
\end{table}

%%==================
\subsubsection{Effect of Monte Carlo Sample Number}
\begin{figure*}[!t]
\centering
\subfloat[]{\includegraphics[width=0.27\textwidth]{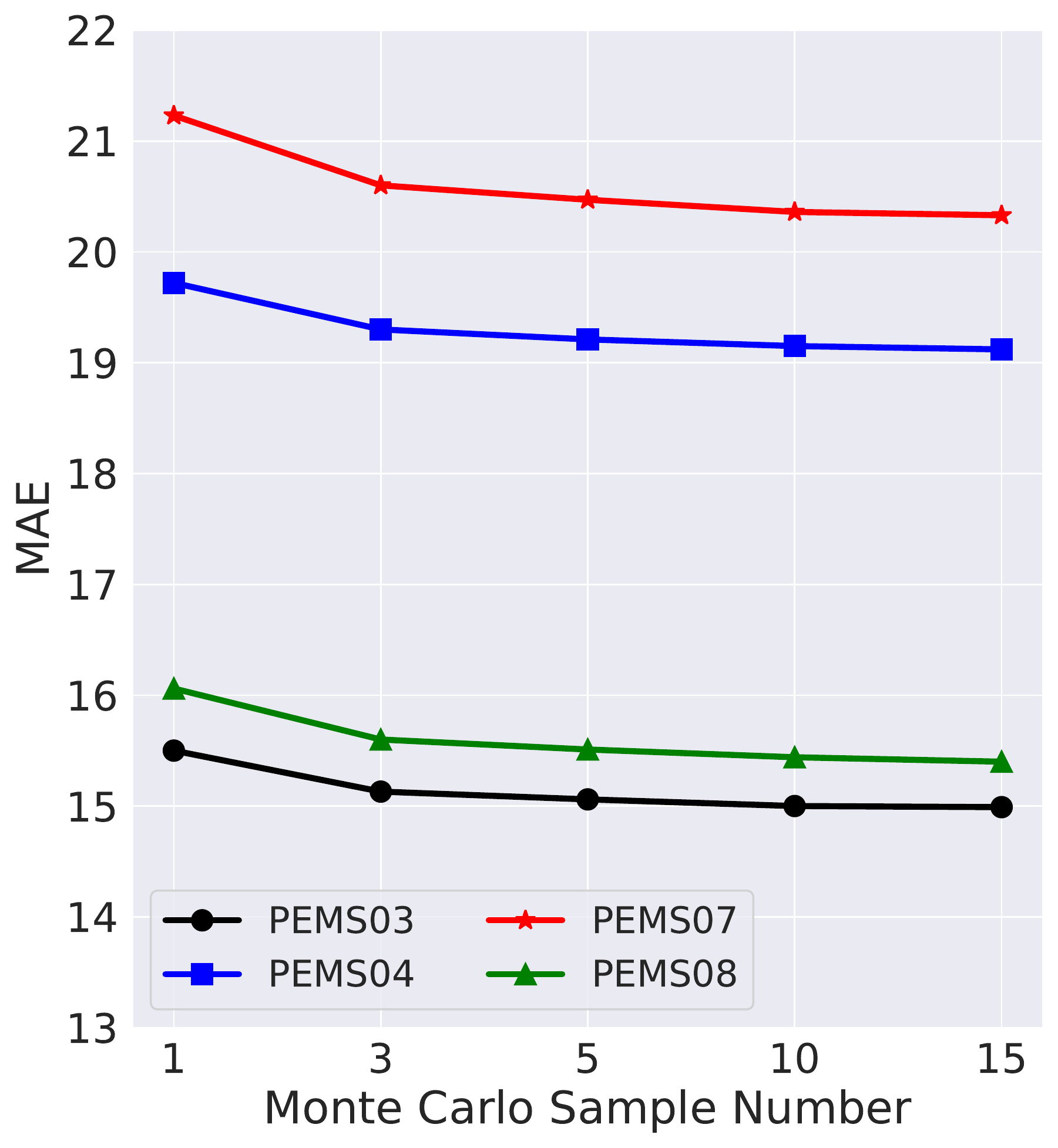}%
\label{fig:mae_mc}}
%~%\\%hfil
\subfloat[]{\includegraphics[width=0.27\textwidth]{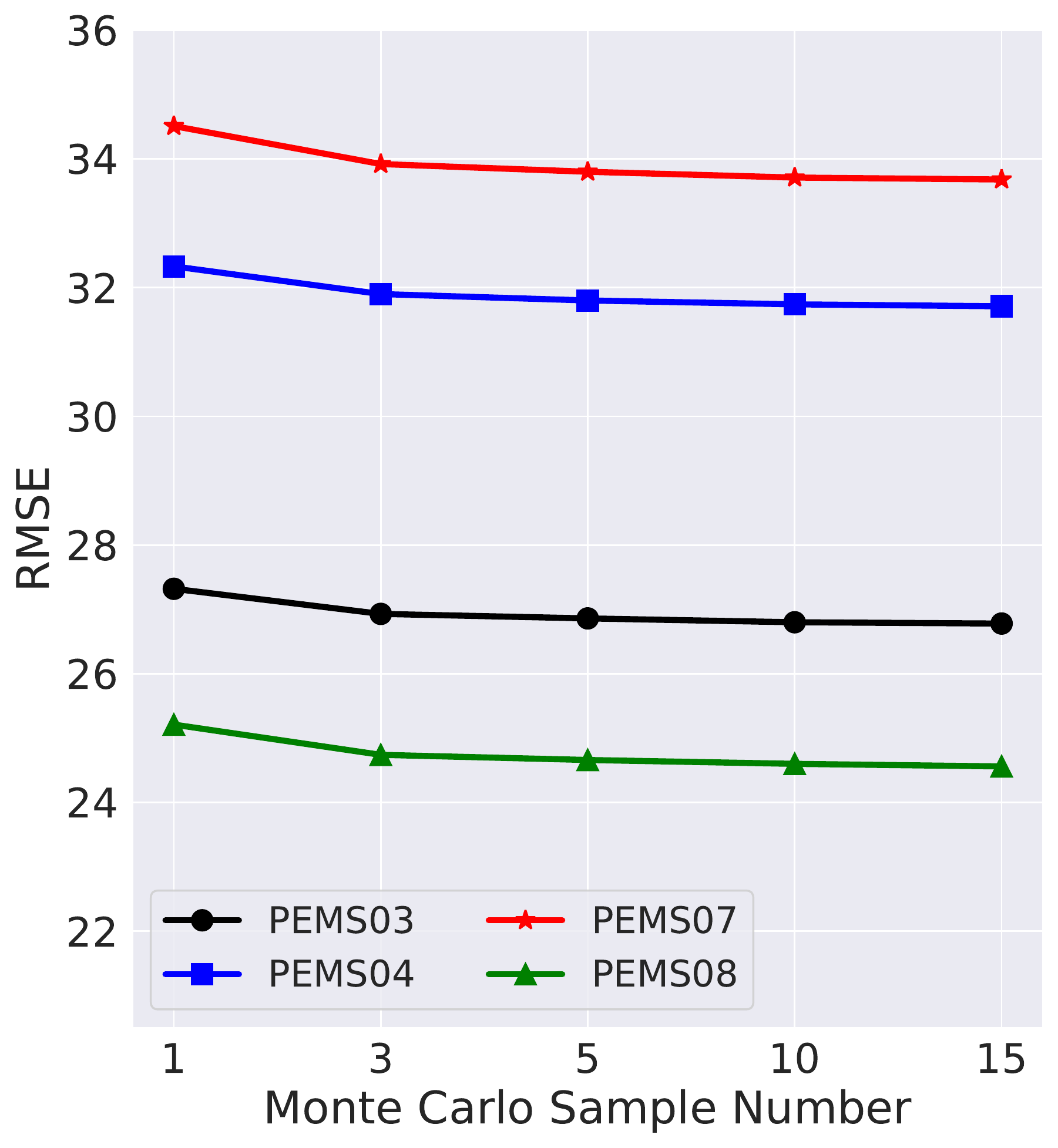}%
\label{fig:rmse_mc}}
%~%\\%hfil
\subfloat[]{\includegraphics[width=0.27\textwidth]{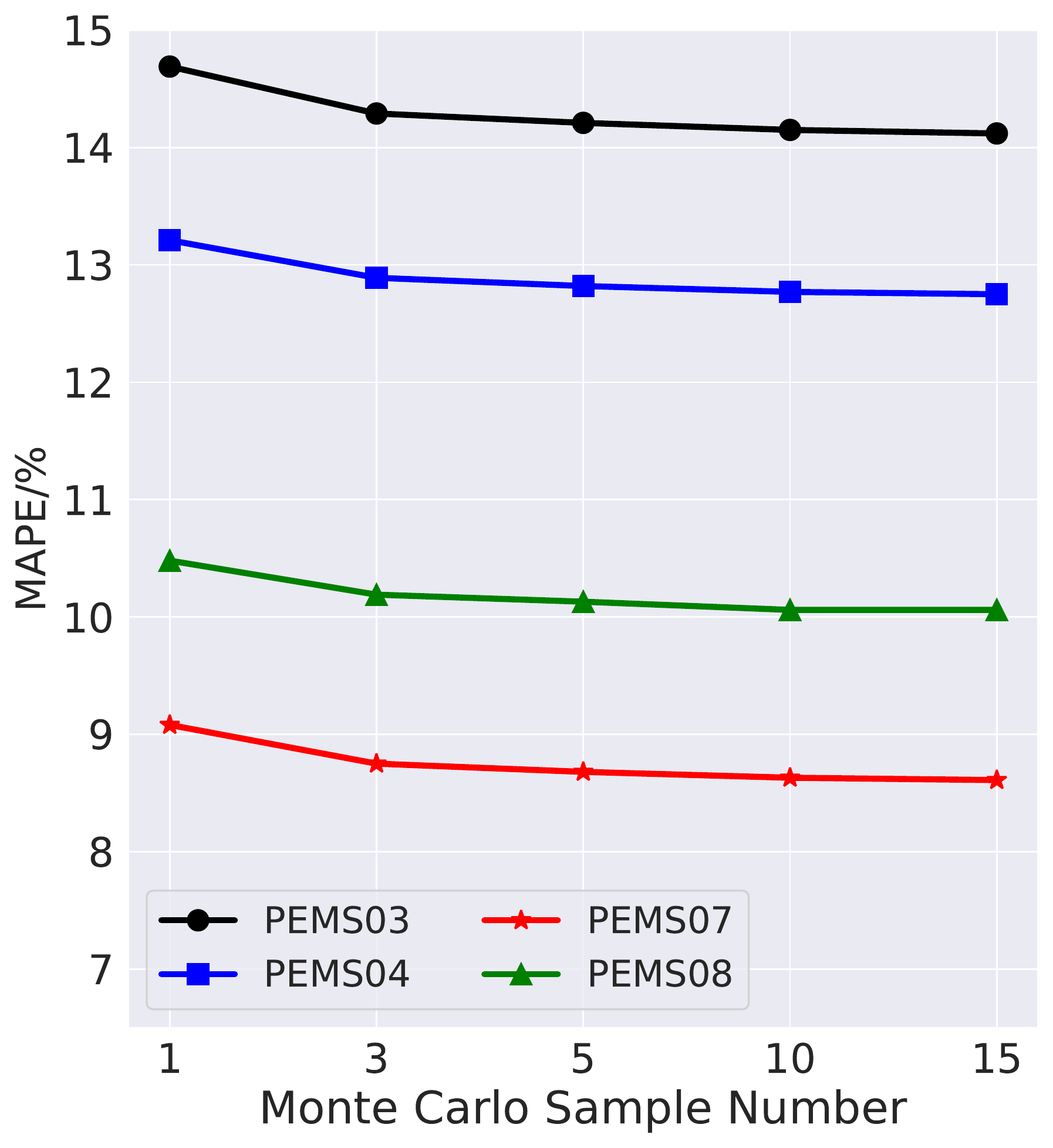}%
\label{fig:mape_mc}}
\caption{Prediction results with respect to different numbers of Monte Carlo samples. (a) MAE. (b) RMSE. (c) MAPE.}
\label{fig:mc_results}
\end{figure*}

To investigate how the number of Monte Carlo samples affects the model performance, the sample number is set to $1$, $3$, $5$, $10$ and $15$.
As shown in Fig. \ref{fig:mc_results}, the performance of the proposed method enhances as the number of Monte Carlo samples rises, and only a small number of Monte Carlo samples are required to provide high prediction performance.
The performance gradually saturates when the sample size approaches $15$.  
%Finally, 
Accordingly, for the trade-off between the model performance and the inference time cost, the test-time sample number can be fixed to $10$ at test time.%\task[(07-04) A bit strange why this sentence appears here.]{}

%\newpage
%%==================
\section{Conclusion}
\label{sec:conclusion}
In this paper, we introduce a novel and unified uncertainty quantification method for traffic forecasting called DeepSTUQ.
The proposed method consists of three components.
1) To model the aleatoric uncertainty, a hybrid loss function is used to train a base spatio-temporal model. 
2) To model the epistemic uncertainty, the merits of variational inference and deep ensembling are combined through the dropout pre-training and AWA re-training.
3) The model is calibrated on the validation dataset using a post-processing calibration method based on Temperature Scaling to further improve the uncertainty estimation performance.   
Four distinct public datasets are then subjected to thorough experiments.
The results indicate that DeepSTUQ outperforms contemporary state-of-the-art spatio-temporal models and uncertainty quantification methods.

Future research may include the incorporation of additional relevant information, e.g., weather forecasts, to enhance the traffic forecasting performance. 
Other deep learning techniques, such as Attention mechanism and hierarchical structures, are also worthy of investigation. 
The temporal and spatial features of traffic data at different scales are a promising future research direction, especially for long-term traffic prediction where multi-modality and seasonality play critical roles.

%%==================
%\newpage
%\pagebreak 
%\nocite{*}
\bibliographystyle{IEEEtran}
\bibliography{IEEEabrv,refs}
\end{document}